\documentclass{article}

\usepackage[preprint]{neurips_2021}

\usepackage[utf8]{inputenc} % allow utf-8 input
\usepackage[T1]{fontenc}    % use 8-bit T1 fonts
\usepackage{hyperref}       % hyperlinks
\usepackage{url}            % simple URL typesetting
\usepackage{booktabs}       % professional-quality tables
\usepackage{amsfonts}       % blackboard math symbols
\usepackage{nicefrac}       % compact symbols for 1/2, etc.
\usepackage{microtype}      % microtypography
\usepackage{xcolor}         % colors

\usepackage{natbib} 
\usepackage{graphicx}
\usepackage{subfigure}
\usepackage{dsfont}
\usepackage{amsmath}
\usepackage{amssymb}
\usepackage{array}
\usepackage{algorithm, algorithmic}
\usepackage{booktabs}
\usepackage{adjustbox}

\def\RR{\mathbb{R}}

\def\bE{\mathbf{E}}
\def\cE{\mathcal{E}}
\def\cH{\mathcal{H}}
\def\cX{\mathcal{X}}
\def\cY{\mathcal{Y}}
\def\cN{\mathcal{N}}

\newcommand{\bphi}[2]{\mathcal{I}_{#1}(\varphi_{#2})}

\def\argmin{\mathrm{argmin}}

\usepackage{amsthm}
\newtheorem{theore}{Theorem}

\newtheorem{lemm}{Lemma}

\theoremstyle{definition}
\newtheorem{definitio}{Definition}

\newtheorem{remar}{Remark}

\title{On Invariance Penalties for Risk Minimization}

\author{
 Kia Khezeli\\
 Mayo Clinic\\
 \texttt{khezeli.kia@mayo.edu}\\
 \And
 Arno Blaas\\
 Oxford University\\
 \texttt{arno@robots.ox.ac.uk}\\
 \And
 Frank Soboczenski\\
 King's College London\\
 \texttt{frank.soboczenski@kcl.ac.uk}\\
 \And
 Nicholas Chia\\
 Mayo Clinic\\
 \texttt{chia.nicholas@mayo.edu}\\
 \And
 John Kalantari\\
 Mayo Clinic\\
 \texttt{kalantari.john@mayo.edu}\\
}

\begin{document}

\maketitle

\begin{abstract}
The Invariant Risk Minimization (IRM) principle was first proposed by \citet{arjovsky2019invariant} to address the domain generalization problem by leveraging data heterogeneity from differing experimental conditions. Specifically, IRM seeks to find a data representation under which an optimal classifier remains invariant across all domains. Despite the conceptual appeal of IRM, the effectiveness of the originally proposed invariance penalty has recently been brought into question. In particular, there exists counterexamples for which that invariance penalty can be arbitrarily small for non-invariant data representations. We propose an alternative invariance penalty by revisiting the Gramian matrix of the data representation. We discuss the role of its eigenvalues in the relationship between the risk and the invariance penalty, and demonstrate that it is ill-conditioned for said counterexamples. The proposed approach is guaranteed to recover an invariant representation for linear settings under mild non-degeneracy conditions. Its effectiveness is substantiated by experiments on DomainBed and InvarianceUnitTest, two extensive test beds for domain generalization.
\end{abstract}

\section{Introduction}
Under the learning paradigm of Empirical Risk Minimization (ERM) \citep{vapnik1992principles}, data is assumed to consist of independent and identically distributed (iid) samples from an underlying generating distribution. As the data generating distribution is often unknown in practice, ERM seeks predictors with minimal average training error (i.e., empirical risk) over the training set. Despite becoming a ubiquitous paradigm in machine learning, a growing body of literature  \citep{arjovsky2019invariant, teney2020unshuffling} has revealed that ERM and the the common practice of shuffling data inadvertently results in capturing all correlations found in the training data, whether spurious or causal,  and produces models that fail to \textit{generalize} to test data. The potential variation of experimental conditions from training to the utilization in real-world applications, manifests in discrepancy between training and testing distributions. This, in turn, highlights the need for machine learning algorithms to \textit{generalize out-of-distribution (OoD)}. 

Shuffling and treating data as iid risks possibly losing important information about the underlying conditions of the data generating process. Instead, partitioning training data into \textit{environments}, e.g., based on the conditions under which data is generated, can exploit these differences to enhance generalization. Based on this observation, \citet{arjovsky2019invariant} introduce the principle of Invariant Risk Minimization (IRM) with the objective of finding a predictor that is invariant across all training environments (see Definition \ref{def:inv pred} and Equation \ref{eq:IRM}). Because of the conceptually appealing nature of IRM and its potential to address the OoD-generalization problem, there is a stream of literature scrutinizing various facets of the original framework, e.g., extensions to other settings including online learning \citep{javed2020learning} and treatment effect estimation \citep{shi2020invariant}, introducing game-theoretic interpretations \citep{ahuja2020invariant}, and raising concerns on the drawbacks and limitations of current IRM implementations \citep{rosenfeld2021the, kamath2021does}. 

For an in-depth overview of the broader generalization literature, we refer the interested reader to \citep{arjovsky2020out} and the references therein, and for an empirical evaluation of the performance of a number of the state-of-the-art methods on various test cases, we refer the reader to \citep{gulrajani2020search}. 

\subsection{Contributions}
In this paper, we propose a novel invariance penalty for practical implementation of IRM. Our proposed invariance penalty is directly related to risk. More precisely, we show that the risk in each environment under an arbitrary classifier equals to the risk under the invariant classifier for that environment plus the proposed invariance penalty between the said classifier and the optimal one. Moreover, we show that the proposed framework finds an invariant predictor for the setting in which the data is generated according to a linear Structural Equation Model (SEM) when provided a sufficient number of training environments under a mild non-degeneracy condition, which is similar in nature to the ones considered in \citep{arjovsky2019invariant, rosenfeld2021the}.

In addition, this work serves to illustrate the importance of the eigenstructure of the Gram matrix of the data representation for IRM. In particular, we show that the Gram matrix is ill-conditioned in the counterexample of \cite{rosenfeld2021the} where the invariance penalty of \cite{arjovsky2019invariant} is made arbitrarily small. Moreover, we characterize the difference between our proposed invariance penalty and the one proposed by \cite{arjovsky2019invariant} in terms of the eigenvalues of the Gram matrix of the data representation. This eigenstructure plays a significant role in the failure of invariance penalties including the one proposed by \cite{arjovsky2019invariant}. 

Finally, we evaluate our method on various test cases including DomainBed \citep{gulrajani2020search}, a test bed including various benchmark data sets for domain generalization, and InvarianceUnitTest \citep{aubinlinear2020}, a test bed with three synthetically generated data sets capturing different structures of spurious correlations. We demonstrate the competitiveness of our proposed framework with other state-of-the-art methods for addressing OoD generalization problem.

\subsection{Organization}
The remainder of the paper is organized as follows. In Section \ref{sec:IRM}, we formally define the notion of invariant prediction, the invariant risk minimization principle, and its relaxation proposed by \citet{arjovsky2019invariant}. In Sections \ref{sec:IRMreg} and \ref{sec:LIRM}, we introduce our more practical implementation and the rationale for its design. In Section \ref{sec:experiments}, we evaluate the efficacy of our proposed model and compare it with other variations of IRM over a series of experiments. We conclude the paper in Section \ref{sec:conclusion}. All mathematical proofs are presented in the Appendix.

\section{Background: Invariant Prediction}\label{sec:IRM}
In this paper, we consider data $(X^e,Y^e)$ collected from multiple training environments $\cE_{\mathrm{tr}}$ where the distribution of $(X^i,Y^i)$ and $(X^j,Y^j)$ may be different for $i\neq j$ with $i,j \in \cE_{\mathrm{tr}}$. We denote by $R_e$ the risk under environment $e$. That is, for predictor $f:\cX\rightarrow \cY$, and loss function $\ell:\cY\times \cY\rightarrow \RR$, the risk under  environment $e$ is defined as
\begin{align}
    R_e(f) = \bE_{X^e,Y^e}\left[\ell(f(X^e),Y^e) \right]. \label{eq:risk}
\end{align}
\subsection{Invariant Risk Minimization}
\citet{arjovsky2019invariant} define the notion of invariant predictors under a multi-environment setting as follows.
\begin{definitio}[Invariant Predictor] \label{def:inv pred}
A data representation $\varphi: \cX\rightarrow \cH$ is said to elicit an invariant predictor $w\circ \varphi$ across environments $\cE$ if there exists a classifier $w: \cH\rightarrow \cY$, which is optimal for all environments, i.e., $w\in \argmin_{\Tilde{w}: \cH\rightarrow \cY}\  R_e\left(\Tilde{w} \circ \varphi\right)$ for all $e\in \cE$.
\end{definitio}
To find such invariant predictors, \citet{arjovsky2019invariant} introduce the notion of the Invariant Risk Minimization (IRM) principle:
\begin{align}
\begin{aligned}
     &\min_{\substack{\varphi: \cX\rightarrow \cH\\w: \cH\rightarrow \cY}} \qquad \sum_{e\in \cE_{\mathrm{tr}}} R_e\left(w \circ \varphi\right) \\
    &\mathrm{subject~to}\quad w\in \underset{\Tilde{w}: \cH\rightarrow \cY}{\mathrm{argmin}}\  R_e\left(\Tilde{w} \circ \varphi\right),\ \forall e\in \cE_{\mathsf{tr}}.
\end{aligned}\label{eq:IRM}
\end{align}
As this bi-leveled optimization problem is rather intractable, \citet{arjovsky2019invariant} propose a practical implementation of IRM by relaxing the invariance constraint (which itself requires solving an optimization problem) to an invariance penalty. We review it in what follows.
\subsection{IRMv1: A Relaxation of IRM}
In order to provide an implementation of IRM, \citet{arjovsky2019invariant} restrict the classifier $w$ to linear functions, i.e., 
\begin{align}
\begin{aligned}
     &\min_{\substack{\varphi: \cX\rightarrow \cH\\ w\in \RR^{d_\varphi}}} \qquad \sum_{e\in \cE_{\mathrm{tr}}} R_e\left(w^\top \varphi\right)  \\
    &\mathrm{subject~to}\quad w \in \underset{\tilde{w}\in \RR^{d_\varphi}}{\argmin}\  R_e\left(\Tilde{w}^\top \varphi\right),\ \forall e\in \cE_{\mathsf{tr}}. 
\end{aligned}\label{eq:IRM 2}
\end{align}
To motivate their proposed penalty, \cite{arjovsky2019invariant} first consider the squared loss, i.e., $\ell(f(x), y) = \|f(x)-y\|^2$ where $\|\cdot\|$ denotes the Euclidean norm. Let the matrix $\bphi{e}{}$ be defined as 
\begin{align}
    \bphi{e}{} := \bE_{X^e}\left[\varphi(X^e)\varphi(X^e)^\top\right]. \label{eq:bphi}
\end{align}
Assuming that $\bphi{e}{}$ is full rank for a fixed $\varphi$, its respective optimal classifier is unique, i.e., 
\begin{align}
    \underset{\tilde{w}\in \RR^{d_\varphi}}{\argmin}\  R_e\left(\Tilde{w}^\top \varphi\right) = w_e^\star(\varphi),
\end{align}
where
\begin{align}
    w_e^\star(\varphi) = \bphi{e}{}^{-1} \bE_{X^e, Y^e}\left[\varphi(X^e)Y^e\right]. \label{eq:LSE}
\end{align}
To relax the constraint $w-w_e^\star(\varphi)=0$ to a penalty, \citet{arjovsky2019invariant} first consider the natural choice of $\|w-w_e^\star(\varphi)\|^2$. However, they show that this penalty does not capture invariance by constructing an example for which $\|w-w_e^\star(\varphi)\|^2$ is not well-behaved (see Section \ref{sec:role eig} for more details). Using the insight from this example, they propose $\| \bphi{e}{}(w - w_e^\star(\varphi)) \|^2$ as an invariant penalty. For the squared loss, one can show that
\begin{align}
   \left\| \bphi{e}{}(w - w_e^\star(\varphi))  \right\|^2 = (1/4) \left\| \nabla_{w} R_e(w^\top \varphi) \right\|^2. \label{eq:penalty equiv}
\end{align}
Hence, their proposed penalty is given by
\begin{align}
    &\rho_e^{\mathrm{IRMv1}}(\varphi, w):= \left\| \nabla_{w} R_e(w^\top \varphi) \right\|^2. \label{eq:Arj pen}
\end{align}
Using the penalty \eqref{eq:Arj pen}, the relaxation of IRM is given by
\begin{align}
    \min_{\varphi,\ w}\  \sum_{e\in \cE_{\mathrm{tr}}}\ R_e(w^\top \varphi) + \lambda \rho_e^{\mathrm{IRMv1}}(\varphi, w), \label{eq:IRM relaxed}
\end{align}
where $\lambda\geq 0$ is the penalty coefficient. Notice that for a given $w$ and  $\varphi$ the predictor $w\circ \varphi$ can be expressed using different classifiers and data representations, i.e., $w\circ \varphi = \Tilde{w} \circ \Tilde{\varphi}$  where $\tilde{w} = w\circ \psi^{-1}$ and $\tilde{\varphi}=\psi\circ \varphi$ for some invertible mapping  $\psi: \cH\rightarrow \cH$. Hence, in principle, it is possible to fix $w$ without loss of generality.  By relying on this observation, \citet{arjovsky2019invariant} fix the classifier as a scalar $w=1$, and, thus, search for invariant data representation of the form $\varphi\in \RR^{1\times d_x}$. Their relaxation of IRM, which they refer to by IRMv1 is given by
\begin{align} \tag{IRMv1}
    &\min_{ \varphi} \ \sum_{e\in \cE_{\mathrm{tr}}} R_e\left(\varphi\right) + \lambda \rho_e^{\mathrm{IRMv1}}(\varphi, 1.0). \label{eq:irmv1}
\end{align}
Although equation \eqref{eq:penalty equiv} only holds for squared loss, \citet[Theorem 4]{arjovsky2019invariant} show that for all differentiable loss functions $(w^\top \Phi)^\top \nabla_w R(w^\top \Phi) = 0$ if and only if $w$ is optimal for all environments. Here, the matrix $\Phi$ parameterizes the data representation. Hence, they justify the choice of $\|\nabla_{w|w=1.0} R_e(w^\top \varphi)\|^2$ as an invariance penalty for other loss functions, e.g., cross-entropy loss. However, recently \citet{rosenfeld2021the} constructed a counterexample by finding a non-invariant data representation for which the penalty $\|\nabla_{w|w=1.0} R_e(w^\top \varphi)\|^2$ with logistic loss is arbitrarily small. 

Recall the assumption of invertability of $\bphi{e}{}$, which was useful in the derivation of the invariance penalty $ \rho_e^{\mathrm{IRMv1}}(\varphi, w)$ for squared loss. In what follows, we investigate the role of the eigenstructure of $\bphi{e}{}$ in relation to invariance penalization, and in particular, the existing counterexamples for the two penalties considered in this section.

\subsection{The Role of $\bphi{e}{}$}\label{sec:role eig}
In proposing their invariance penalty, \citet{arjovsky2019invariant} consider an example in which $\varphi_c(x)$ is parameterized by a variable $c\in \RR$ where $c=0$ for the invariant data representation (see Appendix \ref{app:arj} for further details). Figure \ref{fig:penalties} depicts various candidates for invariance penalty at the invariant classifier $w = w_\mathrm{inv}$. As \citet{arjovsky2019invariant} point out, $\|w_\mathrm{inv} - w^\star_e(\varphi_c)\|^2$ is a poor choice for the invariance penalty as it is discontinuous at the invariant representation with $c=0$, and vanishes as $c\rightarrow\infty$. Interestingly, $\bphi{e}{c}$ is ill-conditioned for both small and large $c$'s. More precisely, it holds that $\lim_{c\rightarrow 0} \kappa(\bphi{e}{c}) = \lim_{c\rightarrow +\infty} \kappa(\bphi{e}{c}) = +\infty$ where $\kappa(\cdot)$ denotes the condition number. That is, for a normal matrix $A$, its condition number is $\kappa(A):= |\lambda_{\max}(A)|/|\lambda_{\min}(A)|$ where $\lambda_{\max}$ and $\lambda_{\min}$ denote its maximum and minimum eigenvalues, respectively. Although multiplying $(w_\mathrm{inv} - w^\star_e(\varphi_c))$ by $\bphi{e}{c}$ circumvents the poor behavior of the invariance penalty for this example, it may not appropriately capture invariance in general as argued by \citet{rosenfeld2021the}.

We now examine the counterexample introduced by \citet{rosenfeld2021the}. They consider a setting in which the data is generated according to a Structural Equation Model (SEM) (see Section \ref{sec:eig revisted}). They show that for this setting, there exists a non-invariant data representation under which $\|\nabla_{w} R_e(w^\top \varphi)\|^2$ with logistic loss is arbitrarily small and hence it is poor discrepancy as an invariance penalty. For their counterexample, the matrix $\bphi{e}{c}$ is also ill-conditioned. 

We provide a detailed derivation of the condition number of $\bphi{e}{c}$ for both \citet{arjovsky2019invariant} and \citet{rosenfeld2021the} in Appendices \ref{app:arj} and \ref{app:rosen}, respectively.

\begin{figure}[h]
    \centering
    \includegraphics[width = 0.75\columnwidth]{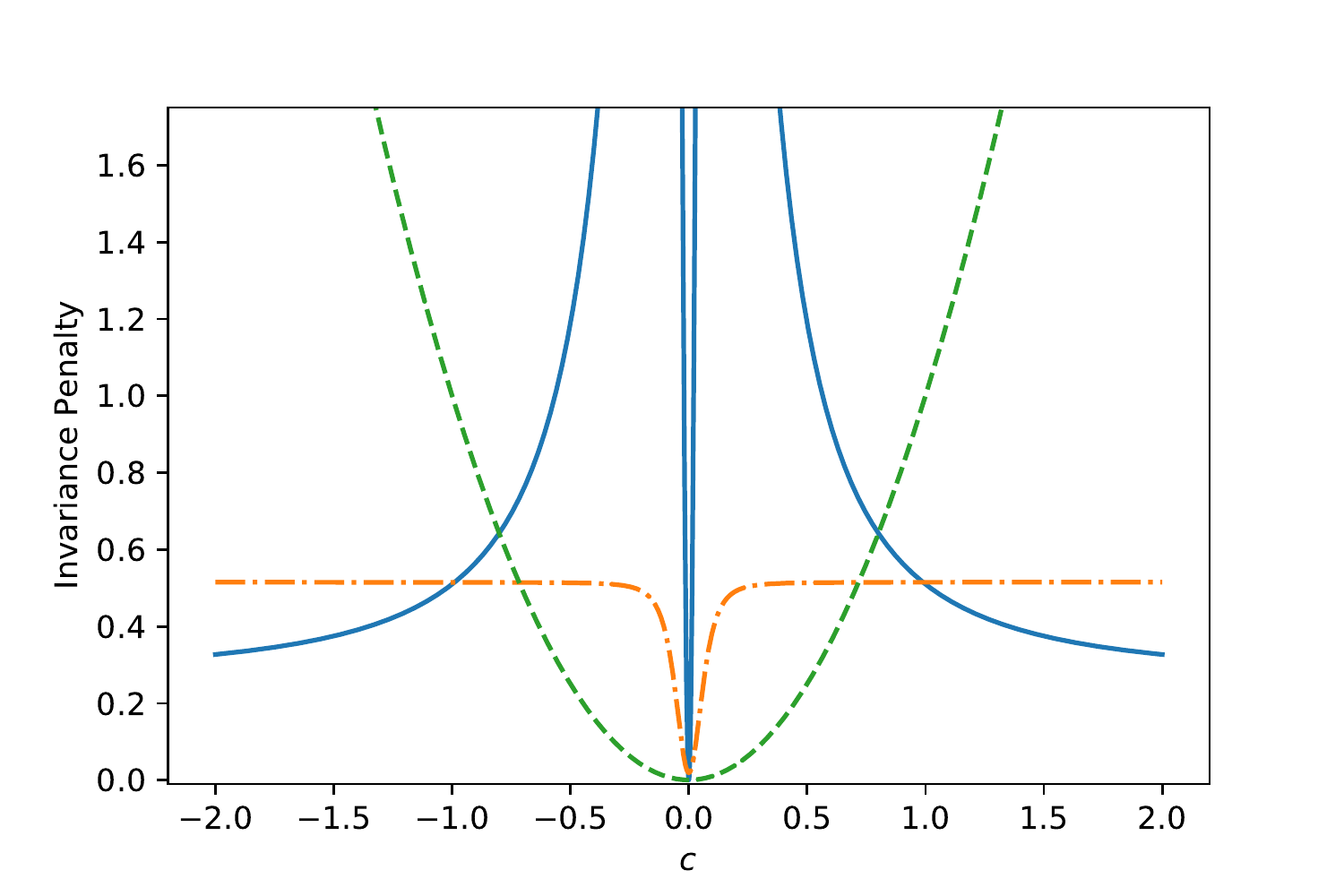}
    \caption{Invariance penalties $\|w_\mathrm{inv} - w^\star_e(\varphi_c)\|^2$, $\|\bphi{e}{}^{1/2}(w_\mathrm{inv} - w^\star_e(\varphi_c))\|^2$, and $\|\bphi{e}{}(w_\mathrm{inv} - w^\star_e(\varphi_c))\|^2$ are depicted in solid blue, dashed-dot orange, and dashed green, respectively. }
    \label{fig:penalties}
\end{figure}
\section{IRMv2: An Alternative Penalty}\label{sec:IRMreg}
As discussed in Section \ref{sec:role eig}, both $\|w - w^\star_e(\varphi_c)\|^2$ and $\|\bphi{e}{}(w - w^\star_e(\varphi_c))\|^2$ may be inappropriate choices for the invariance penalty due to their instability in terms of the eigenstructure of $\bphi{e}{}$. We revisit the structure of the risk in order to propose an alternative penalty. In particular, in the following Lemma, we provide the sub-optimality gap of risk under an arbitrary classifier in comparison to the optimal classifier.
\begin{lemm} \label{lem:risk diff}
Consider squared loss function. Let $w\in \RR^{d_\varphi}$ and $w_e^\star(\varphi)$ as defined in Equation \eqref{eq:LSE}.  Then, 
\begin{align}
     &R_e\left(w^\top \varphi\right) = R_e\left(w_e^\star(\varphi)^\top \varphi\right) + \left\| \bphi{e}{}^{1/2} \left(w - w_e^\star(\varphi)\right)  \right\|^2. \label{eq:risk diff}
\end{align}
\end{lemm}
Based on Lemma \ref{lem:risk diff}, we propose an invariance penalty that is directly comparable to risk. 
\begin{align}
    &\rho_e^{\mathrm{IRMv2}}(\varphi, w) :=  \left\| \bphi{e}{}^{1/2} \left(w - w_e^\star(\varphi)\right)  \right\|^2. \label{eq:LIRM pen}
\end{align}
The relaxation of IRM using the penalty \eqref{eq:LIRM pen} is then given by
\begin{align}
    \min_{\varphi,\ w}\  \sum_{e\in \cE_{\mathrm{tr}}}\ R_e(w^\top \varphi) + \lambda \rho_e^{\mathrm{IRMv2}}(\varphi, w). \label{eq:IRM w}
\end{align}
We further simplify the relaxation \eqref{eq:IRM w} by finding its optimal classifier for a fixed data representation defined as 
\begin{align}
    w^\star(\varphi) &:= \argmin_w \ \sum_{e\in \cE_{\mathrm{tr}}} R_e\left(w^\top \varphi\right) + \lambda \rho_e^{\mathrm{IRMv2}}(\varphi, w).\label{eq:w_opt}
\end{align}
In the following Lemma, we leverage on the structure of the squared loss to find $w^\star(\varphi)$.
\begin{lemm} \label{lem:opt class}
Consider squared loss function and fixed $\varphi$. Let $w_e^\star(\varphi)$ and $w^\star(\varphi)$ as defined in Equations \eqref{eq:LSE} and \eqref{eq:w_opt}, respectively. Then,
\begin{align}
    w^\star(\varphi) = \left(\sum_{e\in \cE_{\mathrm{tr}}} \bphi{e}{}\right)^{-1}\left(\sum_{e\in \cE_{\mathrm{tr}}} \bphi{e}{}w_e^\star(\varphi)\right). \label{eq:w*}
\end{align}
Moreover, 
\begin{align}
     \argmin_w \ \sum_{e\in \cE_{\mathrm{tr}}} R_e\left(w^\top \varphi\right)=w^\star(\varphi). \label{eq:w_erm}
\end{align}
\end{lemm}
Based on Lemmas \ref{lem:risk diff} and \ref{lem:opt class}, we propose the following relaxation of IRM, which we refer to by IRMv2.
\begin{align} \tag{IRMv2}
    &\min_{ \varphi} \ \sum_{e\in \cE_{\mathrm{tr}}} R_e\left( w^\star(\varphi)^\top \varphi\right) + \lambda \rho_e^{\mathrm{IRMv2}}(\varphi,  w^\star(\varphi)). \label{eq:irmv2}
\end{align}
We provide the pseudo-code for IRMv2 in Algorithm \ref{alg:IRMv2}.
\begin{algorithm}[h]
\caption{IRMv2}\label{alg:IRMv2}
\vspace{1ex}
\begin{algorithmic}[1]
\STATE \textbf{Input:} Data set: $D_e$ for $e\in \cE_{\mathrm{tr}}$. Loss function: Squared loss, Parameters: penalty coefficient $\lambda\geq 0$, data representation parameters $\theta\in \RR^{d_\theta}$, learning rate $\eta_t$, training horizon $T$. \\[0.5em]
\STATE \textbf{Initialize} $\theta_1$ randomly \\[0.5em]
\STATE \textbf{for $t=1,2,\ldots, T $ do}
\STATE \qquad \textbf{for $e\in \cE_{\mathrm{tr}} $ do}
\STATE \qquad\qquad compute the LSE $w^\star_e(\varphi_{\theta_{t}})$ according to Eq. \eqref{eq:LSE}\\[0.5em]
\STATE \qquad compute the optimal classifier $w^\star(\varphi_{\theta_{t}})$ according to Eq. \eqref{eq:w_opt}\\[0.5em]
\STATE \qquad $\mathcal{L}_t(\varphi_{\theta_{t}})\leftarrow \sum_{e\in \cE_{\mathrm{tr}}} \mathcal{R}_e(w^\star(\varphi_{\theta_{t}})^\top\varphi_{\theta_{t}}) + \lambda \rho_e^{\mathrm{IRMv2}}(\varphi_{\theta_{t}}, w^\star(\varphi_{\theta_{t}}))$\\[0.5em]
\STATE \qquad $ \theta_{t+1} \leftarrow \theta_t - \eta_t \nabla_{\theta_{t}} \mathcal{L}_t(\varphi_{\theta_{t}})$\\[0.5em]
\STATE \textbf{Output} predictor $w^\star(\varphi_{\theta_{T}})^\top \varphi_{\theta_{T}}$.
\end{algorithmic}
\end{algorithm}

There are several factors distinguishing IRMv2 from IRMv1. First, IRMv2 relies on the optimal classifier $w^\star(\varphi)$ while $w=1.0$ in IRMv1. Second, the loss function in IRMv2 is squared loss while IRMv1 allows for utilization of other loss functions. Although this additional flexibility of IRMv1 may seem appealing, the counterexample of \citet{rosenfeld2021the} shows the failure of the penalty of IRMv1 to capture invariance for logistic loss. Finally, $\bphi{e}{}$ is incorporated differently in the invariance penalty of IRMv1 and IRMv2. We formalize this latter observation in the following Section.

\subsection{IRMv1A: An Adaptive Penalty Coefficient}
We first bound the invariance penalty of IRMv1 in terms of the penalty of IRMv2 and the eigenvalues of $\mathcal{I}_e(\varphi)$. Then, based on this comparison, we propose an adaptive approach in choosing the penalty coefficient for IRMv1, which we refer to as IRMv1-Adaptive (IRMv1A).
\begin{lemm} \label{lem:adaptive irm}
Let $\rho_e^{\mathrm{IRMv1}}(\varphi, w)$ and $\rho_e^{\mathrm{IRMv2}}(\varphi, w)$ be the invariance penalties of the IRMv1 and IRMv2 defined in Equations \eqref{eq:Arj pen} and \eqref{eq:LIRM pen}, respectively. Then,
\begin{align}
    \lambda_{\min}\left(\mathcal{I}_e(\varphi)\right)\rho_e^{\mathrm{IRMv2}}(\varphi, w) \leq \rho_e^{\mathrm{IRMv1}}\left(\varphi, w\right) \leq \lambda_{\max}\left(\mathcal{I}_e(\varphi)\right)\rho_e^{\mathrm{IRMv2}}(\varphi, w).
\end{align}
\end{lemm}
The proof of Lemma \ref{lem:adaptive irm} directly follows from the definition of the invariance penalties $\rho_e^{\mathrm{IRMv1}}(\varphi, w)$ and $\rho_e^{\mathrm{IRMv2}}(\varphi,w)$, and the fact that for a symmetric matrix $A\in \RR^{d\times d}$ and a vector $u\in \RR^d$, it holds that $ \lambda_{\min}(A) \|u\|^2 \leq  u^\top A u\leq \lambda_{\max}(A) \|u\|^2$.

Using Lemma \ref{lem:adaptive irm}, we suggest the following rule for the penalty coefficient of IRMv1.
\begin{align}
    \lambda_e :=  \frac{1}{\lambda_0 + \lambda_{\min}(\mathcal{I}_e(\varphi))} \label{eq:lambda_e}
\end{align}
for a user-specified $\lambda_0\geq 0$. Note that this is an adaptive rule, as $\varphi$ may change throughout training.
\section{Theoretical Results}\label{sec:LIRM}
In this section, we consider the setting introduced by \citet{rosenfeld2021the} and provide theoretical guarantees that IRM with linear classifier and squared loss, and subsequently  IRMv2 recover an invariant predictor. 
\subsection{Problem Setup}\label{sec:sem setup}
We consider a setting in which the data is generated according to a Structural Equation Model \citep{pearl2009causality}. More precisely, for each environment $e$,  $(X^e,Y^e)$ is generated as
\begin{align}
   X^e= S\begin{bmatrix}
    Z_c\\
    Z_e
    \end{bmatrix}, \quad  Y^e=\begin{cases}
    1,& \text{with prob. } \eta,\\
    -1,& \text{with prob. } 1-\eta,
    \end{cases}, \label{eq:data sem}
\end{align}
where $\eta\in [0,1]$, and $S\in \RR^{d\times (d_e+d_c)}$ is a left invertible matrix, i.e., there exists $S^\dag$ such that $S^\dag S = I$. In this model, $Z_c$ captures the causal variables that are invariant across environments, and $Z_e$ captures the spurious environment dependent variables. 

The variables $Z_c$ and $Z_e$ are generated as follows
\begin{align}
    Z_c &= \mu_c Y+ W_c\ \text{where}\ W_c\sim \cN(0, \sigma_c^2I),\\
    Z_e &= \mu_e Y+W_e\ \text{where}\ W_e\sim \cN(0, \sigma_e^2I).
\end{align}
Here, $\mu_c\in \RR^{d_c}$, $\mu_e\in \RR^{d_e}$, and $\cN(\mu, \Sigma)$ denotes multi-variate Gaussian distribution with mean equal to $\mu$ and covariance matrix equal to $\Sigma$. We further assume that $W_c$, $W_e$, and $Y^e$ are independent for all environments.
\subsection{Invariant Representation under IRM}
For the setting introduced in \ref{sec:sem setup}, the invariant data representation is linear. In particular, for any $d\geq d_c$, $\varphi(X^e) = \Phi_d X^e = Z_c$ is an invariant data representation, where
\begin{align}
    \Phi_d := \begin{bmatrix}
    I_{d_c\times d_c}& \mathbf{0}_{d_e\times d_c}\\
    \mathbf{0}_{d_c\times (d-d_c)} & \mathbf{0}_{d_e\times (d-d_c)}
    \end{bmatrix} S^\dag.
\end{align}

Naturally, the possibility of finding an invariant predictor depends on the number and the diversity of training environments. We now introduce non-degeneracy conditions on the training environment under which IRM is guaranteed to find an invariant predictor, provided sufficient number of training environments.

Let $|\cE_{\mathrm{tr}}|>d_e$. As $\mbox{span}(\{\mu_e\}_{e\in \cE_{\mathrm{tr}}})\leq d_e$, for each $e\in \cE_{\mathrm{tr}}$ there exists a set of coefficients $\alpha_i^e$ for $i\in \cE_{\mathrm{tr}}\backslash e$ such that
\begin{align}
    \mu_e = \sum_{i\in \cE_{\mathrm{tr}}\backslash e} \alpha_i^e \mu_i. \label{eq:nondeg mu}
\end{align}
We say that $\cE_{\mathrm{tr}}$ is a \textit{non-degenerate set of environments} if for all $e\in \cE_{\mathrm{tr}}$ it holds that
\begin{align}
 \sum_{i\in \cE_{\mathrm{tr}}\backslash e} \alpha_i^e &\neq 1, \label{eq:nondeg alpha}\\
\mbox{rank}\left(  \Gamma_e\right) &=d_e,\label{eq:nondeg sig}
\end{align}
where $\Gamma_e$ is defined as
\begin{align*}
    \Gamma_e:=\frac{1}{1-\sum_{i\in \cE_{\mathrm{tr}}\backslash e} \alpha_{i}^e } \left(\sigma_e^2 I + \mu_e\mu_e^\top -\sum_{i\in \cE_{\mathrm{tr}}\backslash e} (\sigma_i^2I + \mu_i\mu_i^\top) \alpha_{i}^e\right).
\end{align*}

The conditions \eqref{eq:nondeg alpha} and \eqref{eq:nondeg sig} specify that the span of covariance matrices of $Z_e$ is $R^{d_e}$. This is a natural requirement to eliminate the degrees of freedom on the dependency of the data representation on the enivornment dependent features. We note that the non-degeneracy conditions considered in \cite{rosenfeld2021the} are similar to \eqref{eq:nondeg alpha} and \eqref{eq:nondeg sig} with the difference that instead of depending on covariance matrices of $Z_e$ as in $\eqref{eq:nondeg sig}$, their assumption relies on the variances $\sigma_e^2$. This difference in the non-degeneracy requirements is due to the fact that they consider logistic loss and we consider squared loss.

\begin{theore}\label{thm:lin}
Assume that $|\cE_{\mathrm{tr}}|>d_e$ where $(X^e,Y^e)$ generated according to \eqref{eq:data sem}. Consider a linear data representation $\Phi X = AZ_c + BZ_e$ and a classifier $w(\Phi)$ on top of $\Phi$ that is invariant, i.e., $w(\Phi)=w_e^\star(\Phi)$ for all $e\in  \cE_{\mathrm{tr}}$. If non-degeneracy conditions Eqs. (\ref{eq:nondeg mu}-\ref{eq:nondeg sig}) holds, then either $w(\Phi)=0$ or $B=0$.
\end{theore}

\subsection{Eigenstructure of $\bphi{e}{}$} \label{sec:eig revisted}
In this section, we compare the penalties of IRMv1 and IRMv2 for the counterexample of \cite{rosenfeld2021the}. They consider a data representation $\varphi_\epsilon$ where $\epsilon>1$ determines the extent to which $\varphi_\epsilon(X^e)$ depends on $Z_e$. More specifically, $\varphi_\epsilon$ is defined as
\begin{align}
    \varphi_\epsilon(X^e):=\begin{bmatrix}Z_c\\0\end{bmatrix} + \begin{bmatrix}0\\Z_e\end{bmatrix} \mathds{1}_{ \{Z_e \notin \mathcal{Z}_{\epsilon}\} },
\end{align}
where $\{Z_e\notin \mathcal{Z}_{\epsilon} \}$ is an event with $\mathbf{P}(Z_e\in \mathcal{Z}_{\epsilon}) \leq p_{e,\epsilon}$ where $ p_{e,\epsilon} := \exp(-d_e\min\{\epsilon-1, (\epsilon-1)^2\}/8)$. They show that the invariance penalty of IRMv1 decays at a rate faster than $p_{e,\epsilon}^2$ as $\epsilon$ grows. Thus, the penalty may be arbitrarily small for a large enough $\epsilon$. 

An invariant data representation for this setting is $\varphi_\epsilon(X^e)$ with $\epsilon=1$. Moreover, in Appendix \ref{app:rosen}, we show that $\kappa(\bphi{e}{})\geq c/ p_{e,\epsilon}$ for some constant $c$ that is independent of $\epsilon$. Thus, $\bphi{e}{}$ is ill-conditioned when the penalty of IRMv1 is small. 
\section{Experiments}\label{sec:experiments}
In this section, we empirically evaluate the efficacy of our proposed implementations of IRM, namely, IRMv2 and IRMv1A with IRMv1. We demonstrate the competitiveness of our approach on \textit{InvarianceUnitTests} \citep{aubinlinear2020} and \textit{DomainBed} \citep{gulrajani2020search}, two recently proposed test beds for evaluation of domain generalization methods. In particular, we show that our approach generalizes in one of the InvarianceUnitTests where all other methods failed (i.e., exhibited tests accuracies that are comparable to random guessing). 

\subsection{InvarianceUnitTests}
In this section, we evaluate the efficacy of our proposed approaches for invariance discovery on the \textit{InvarianceUnitTests} recently proposed by \citet{aubinlinear2020}. These unit-tests entail three classes of low-dimensional linear problems, each capturing a different structure for inducing spurious correlations. For completeness, we provide a brief overview of these InvarianceUnitTests before providing a performance comparison between IRMv2, IRMv1A, IRMv1, ERM, Inter-environmental Gradient Alignment (IGA) \citep{koyama2020out}, and AND-Mask \citep{parascandolo2020learning}. The IGA method seeks to elicit invariant predictors by an invariance penalty in terms of the variance of the risk under different environments. The AND-Mask method, at each step of the training process, updates the model using the direction where gradient (of the loss) signs agree across environments.

The data set for each problem falls within the multi-environment setting described in Section \ref{sec:IRM} with $n_e=10^4$. For all problems, the input $x^e\in \RR^d$ is constructed as $x^e = (x_{\mathrm{inv}}^e, x_{\mathrm{spu}}^e)$ where $x_{\mathrm{inv}}^e \in \RR^{d_{\mathrm{inv}}} $ and $x_{\mathrm{spu}}^e\in \RR^{d_{\mathrm{spu}}} $ denote the invariant and the spurious features, respectively. To make the problems more realistic, \citet{aubinlinear2020} repeat each experiment and \textit{scramble} the inputs by multiplying $x^e$ by a rotation matrix. In each problem, the spurious correlations that exist in the training environments are discarded in the test environment by random shuffling. As a basis for comparison, similar to \citet{aubinlinear2020}, we implement an Oracle ERM where the spurious correlations are shuffled in the training data sets as well, and hence, ERM can readily identify them.

Example 1 considers a \textit{regression problem} based on Structural Equation Models \cite{pearl2009causality} where the target variable is a linear function of the invariant variables and the spurious variables are linear functions of the target variable. Example 2 considers a \textit{classification problem} inspired by the infamous cow vs. camel example \cite{beery2018recognition} where spurious correlations are interpreted as background color. Example 3 is based on a classification experiment in \cite{parascandolo2020learning} where the spurious correlations provide a \textit{shortcut} in minimizing the training error while the invariant classifier takes a more complex form.

We summarize the test errors of all methods on the three examples and their scrambled variations in Table \ref{tab:acc unit test}. We observe that on these structured unit-tests, most non-ERM methods are only successful in eliciting an invariant predictor in the linear regression case (Example 1). In particular, other than IRMv2 on Example 2 and IRMv1 on Example 3, all methods fail on these cases, i.e., exhibit test errors comparable to random guessing. As the structure of the spurious correlation is different in each of these examples, these mixed results highlight the challenge of constructing methods that generalize well with minimal reliance  on the underlying causal structure. 

\begin{table}
\begin{center}
\adjustbox{max width=\textwidth}{%
\begin{tabular}{lccccccc}
\toprule
                & ANDMask         & ERM             & IGA             & IRMv1A            & IRMv2         & IRMv1           & Oracle          \\
\midrule
Example1.E0     & 0.07 $\pm$ 0.01 & 1.52 $\pm$ 0.50 & 6.61 $\pm$ 2.77 & 0.07 $\pm$ 0.01 & 0.06 $\pm$ 0.01 & 0.11 $\pm$ 0.02 & 0.05 $\pm$ 0.00 \\
Example1.E1     & 11.54 $\pm$ 0.31 & 14.10 $\pm$ 1.33 & 21.61 $\pm$ 3.85 & 13.53 $\pm$ 1.29 & 13.15 $\pm$ 1.29 & 12.46 $\pm$ 1.25 & 11.26 $\pm$ 0.17 \\
Example1.E2     & 20.56 $\pm$ 0.66 & 23.98 $\pm$ 2.21 & 33.43 $\pm$ 5.09 & 24.11 $\pm$ 2.35 & 23.53 $\pm$ 2.26 & 22.18 $\pm$ 2.27 & 19.99 $\pm$ 0.29 \\
Example1s.E0    & 0.07 $\pm$ 0.01 & 1.66 $\pm$ 0.49 & 6.49 $\pm$ 3.08 & 0.07 $\pm$ 0.01 & 0.06 $\pm$ 0.01 & 0.10 $\pm$ 0.02 & 0.05 $\pm$ 0.00 \\
Example1s.E1    & 12.71 $\pm$ 0.93 & 14.49 $\pm$ 1.45 & 21.83 $\pm$ 4.78 & 13.31 $\pm$ 1.34 & 13.36 $\pm$ 1.28 & 12.61 $\pm$ 1.20 & 11.25 $\pm$ 0.19 \\
Example1s.E2    & 22.59 $\pm$ 1.77 & 24.59 $\pm$ 2.40 & 34.01 $\pm$ 6.55 & 23.68 $\pm$ 2.35 & 23.73 $\pm$ 2.21 & 22.40 $\pm$ 2.16 & 20.01 $\pm$ 0.34 \\
Example2.E0     & 0.43 $\pm$ 0.01 & 0.43 $\pm$ 0.00 & 0.43 $\pm$ 0.00 & 0.43 $\pm$ 0.01 & 0.45 $\pm$ 0.01 & 0.47 $\pm$ 0.01 & 0.42 $\pm$ 0.01 \\
Example2.E1     & 0.50 $\pm$ 0.00 & 0.50 $\pm$ 0.00 & 0.50 $\pm$ 0.00 & 0.50 $\pm$ 0.01 & 0.35 $\pm$ 0.01 & 0.52 $\pm$ 0.01 & 0.50 $\pm$ 0.01 \\
Example2.E2     & 0.42 $\pm$ 0.01 & 0.42 $\pm$ 0.01 & 0.42 $\pm$ 0.01 & 0.42 $\pm$ 0.01 & 0.23 $\pm$ 0.01 & 0.47 $\pm$ 0.01 & 0.42 $\pm$ 0.01 \\
Example2s.E0    & 0.43 $\pm$ 0.01 & 0.43 $\pm$ 0.01 & 0.43 $\pm$ 0.01 & 0.43 $\pm$ 0.01 & 0.51 $\pm$ 0.07 & 0.44 $\pm$ 0.02 & 0.41 $\pm$ 0.02 \\
Example2s.E1    & 0.50 $\pm$ 0.00 & 0.50 $\pm$ 0.00 & 0.50 $\pm$ 0.00 & 0.49 $\pm$ 0.01 & 0.41 $\pm$ 0.06 & 0.50 $\pm$ 0.01 & 0.48 $\pm$ 0.02 \\
Example2s.E2    & 0.42 $\pm$ 0.00 & 0.42 $\pm$ 0.01 & 0.42 $\pm$ 0.01 & 0.42 $\pm$ 0.01 & 0.27 $\pm$ 0.04 & 0.44 $\pm$ 0.02 & 0.41 $\pm$ 0.02 \\
Example3.E0     & 0.41 $\pm$ 0.13 & 0.48 $\pm$ 0.07 & 0.45 $\pm$ 0.12 & 0.35 $\pm$ 0.11 & 0.48 $\pm$ 0.09 & 0.22 $\pm$ 0.13 & 0.01 $\pm$ 0.00 \\
Example3.E1     & 0.44 $\pm$ 0.12 & 0.49 $\pm$ 0.05 & 0.48 $\pm$ 0.07 & 0.35 $\pm$ 0.11 & 0.47 $\pm$ 0.10 & 0.22 $\pm$ 0.13 & 0.01 $\pm$ 0.00 \\
Example3.E2     & 0.42 $\pm$ 0.14 & 0.47 $\pm$ 0.10 & 0.47 $\pm$ 0.09 & 0.35 $\pm$ 0.11 & 0.47 $\pm$ 0.09 & 0.22 $\pm$ 0.13 & 0.01 $\pm$ 0.00 \\
Example3s.E0    & 0.49 $\pm$ 0.05 & 0.49 $\pm$ 0.06 & 0.50 $\pm$ 0.02 & 0.57 $\pm$ 0.18 & 0.49 $\pm$ 0.06 & 0.44 $\pm$ 0.16 & 0.03 $\pm$ 0.01 \\
Example3s.E1    & 0.50 $\pm$ 0.04 & 0.50 $\pm$ 0.00 & 0.50 $\pm$ 0.02 & 0.57 $\pm$ 0.18 & 0.49 $\pm$ 0.07 & 0.44 $\pm$ 0.16 & 0.03 $\pm$ 0.01 \\
Example3s.E2    & 0.49 $\pm$ 0.05 & 0.48 $\pm$ 0.09 & 0.49 $\pm$ 0.07 & 0.57 $\pm$ 0.18 & 0.48 $\pm$ 0.07 & 0.44 $\pm$ 0.17 & 0.03 $\pm$ 0.01 \\
\bottomrule
\end{tabular}}
\end{center}
\caption{Test errors for all algorithms and examples with $(d_{\mathrm{inv}}, d_{\mathrm{spu}}, d_{\mathrm{env}}) = (5, 5, 3)$. The errors for Examples 1.E0 through 1s.E2 are in MSE and all others are classification error. The empirical mean and the standard deviation are computed using 10 independent experiments. An `s' indicates the scrambled variation of its corresponding problem setting, e.g. Example 1s is the scrambled variation of the Example 1 regression setting. }\label{tab:acc unit test} 
\end{table}

\subsection{DomainBed}
DomainBed is an extensive framework released by \cite{gulrajani2020search} to test domain generalization algorithms for image classification tasks on various benchmark data sets. In a series of experiments, \cite{gulrajani2020search} show that enabled by data augmentation various state-of-the-art generalization methods perform similar to each other and ERM on several benchmark data sets. 

Although the integration of additional data sets and algorithms to DomainBed is straightforward, we note that performing an extensive set of experiments requires significant computational resources as also pointed out by \cite{krueger2020out} (see the supplementary material for further details on the computational details). For this reason, we limit the scope of our experiments to the comparison of ERM, IRMv1, IRMv1A, and IRMv2. 

Similar to \cite{gulrajani2020search}, we observe that no method significantly outperforms others on any of the benchmark data sets (see Table \ref{tab:domainbed}). For a complete set of results on DomainBed with various model selection methods, we refer the reader to Appendix \ref{sec:domainbed}.  As these data sets are image based and equipped with data augmentation, they may not provide comprehensive insight on the strengths and weaknesses of domain generalization algorithms on other modes of data, e.g., gathered in real-world applications.

\begin{table}[]
\begin{center}
\adjustbox{max width=\textwidth}{%
\begin{tabular}{lccccc}
\toprule
\textbf{Algorithm}        & \textbf{ColoredMNIST}     & \textbf{RotatedMNIST}  & \textbf{PACS} & \textbf{VLCS}    & \textbf{Avg}              \\
\midrule
ERM                       & 51.7 $\pm$ 0.1            & 96.7 $\pm$ 0.0     & 81.1 $\pm$ 0.1  &    78.8 $\pm$ 0.4     & 77.0                      \\
IRMv1                       & 51.8 $\pm$ 0.2            & 95.2 $\pm$ 0.4  & 78.6 $\pm$ 1.0   &  76.0 $\pm$ 0.5        & 75.4                      \\
IRMv1A                      & 50.9 $\pm$ 0.1            & 64.7 $\pm$ 20.1     &  80.9 $\pm$ 0.0 & 77.3 $\pm$ 0.2      & 68.4                     \\
IRMv2                   & 50.8 $\pm$ 0.4            & 97.1 $\pm$ 0.0     &  82.6 $\pm$ 0.9 &    76.5 $\pm$ 0.4     & 76.8                      \\
\bottomrule
\end{tabular}}
\caption{The test accuracy of ERM and different implementations of IRM on benchmark data-sets. Model selection of the DomainBed is chosen as training-domain validation set. } \label{tab:domainbed}
\end{center}
\end{table}

\section{Conclusion}\label{sec:conclusion}
In this paper, we have presented IRMv2, an alternative implementation of the IRM principle that aims to enable out-of-distribution generalization by finding environment invariant predictors. We establish theoretical results on the effectiveness of our approach in the linear setting. In doing so, we bring forward the importance of the eigenstructure of the Gramian matrix of the data representation. In particular, we show that for the existing counterexample on the potential failure of IRMv1, the aforementioned matrix is ill-conditioned for the invariant representation. This highlights the significance of the span of the data representations in relation to the span of the underlying true invariant features of the data. That is, if the data representation allows for more degrees of freedom than needed to capture invariance, the Gramian matrix of the invariant representation would be ill-conditioned. This observation provides intuition on the underlying reasons why current implementations of IRM may fail. While this work attempts to addresses some of the limitations of IRM that impede its widespread adoption, it leaves for future work subsequent investigations on data gathered from real-world applications beyond curated benchmark data sets.

\bibliographystyle{plainnat}
\bibliography{references.bib}

\begin{thebibliography}{16}
\providecommand{\natexlab}[1]{#1}
\providecommand{\url}[1]{\texttt{#1}}
\expandafter\ifx\csname urlstyle\endcsname\relax
  \providecommand{\doi}[1]{doi: #1}\else
  \providecommand{\doi}{doi: \begingroup \urlstyle{rm}\Url}\fi

\bibitem[Ahuja et~al.(2020)Ahuja, Shanmugam, Varshney, and
  Dhurandhar]{ahuja2020invariant}
Kartik Ahuja, Karthikeyan Shanmugam, Kush Varshney, and Amit Dhurandhar.
\newblock Invariant risk minimization games.
\newblock In \emph{Proceedings of the 37th International Conference on Machine
  Learning}, pages 145--155, 2020.

\bibitem[Arjovsky(2020)]{arjovsky2020out}
Martin Arjovsky.
\newblock \emph{Out of Distribution Generalization in Machine Learning}.
\newblock PhD thesis, New York University, 2020.

\bibitem[Arjovsky et~al.(2019)Arjovsky, Bottou, Gulrajani, and
  Lopez-Paz]{arjovsky2019invariant}
Martin Arjovsky, L{\'e}on Bottou, Ishaan Gulrajani, and David Lopez-Paz.
\newblock Invariant risk minimization.
\newblock \emph{arXiv preprint arXiv:1907.02893}, 2019.

\bibitem[Aubin et~al.(2020)Aubin, Arjovsky, Bottou, and
  Lopez-Paz]{aubinlinear2020}
Benjamin Aubin, Martin Arjovsky, Leon Bottou, and David Lopez-Paz.
\newblock Linear unit tests for invariance discovery.
\newblock In \emph{Causal Discovery and Causality-Inspired Machine Learning
  Workshop at NeurIPS}, 2020.

\bibitem[Beery et~al.(2018)Beery, Van~Horn, and Perona]{beery2018recognition}
Sara Beery, Grant Van~Horn, and Pietro Perona.
\newblock Recognition in terra incognita.
\newblock In \emph{Proceedings of the European Conference on Computer Vision
  (ECCV)}, pages 456--473, 2018.

\bibitem[Gulrajani and Lopez-Paz(2020)]{gulrajani2020search}
Ishaan Gulrajani and David Lopez-Paz.
\newblock In search of lost domain generalization.
\newblock \emph{arXiv preprint arXiv:2007.01434}, 2020.

\bibitem[Javed et~al.(2020)Javed, White, and Bengio]{javed2020learning}
Khurram Javed, Martha White, and Yoshua Bengio.
\newblock Learning causal models online.
\newblock \emph{arXiv preprint arXiv:2006.07461}, 2020.

\bibitem[Kamath et~al.(2021)Kamath, Tangella, Sutherland, and
  Srebro]{kamath2021does}
Pritish Kamath, Akilesh Tangella, Danica~J Sutherland, and Nathan Srebro.
\newblock Does invariant risk minimization capture invariance?
\newblock \emph{arXiv preprint arXiv:2101.01134}, 2021.

\bibitem[Koyama and Yamaguchi(2020)]{koyama2020out}
Masanori Koyama and Shoichiro Yamaguchi.
\newblock Out-of-distribution generalization with maximal invariant predictor.
\newblock \emph{arXiv preprint arXiv:2008.01883}, 2020.

\bibitem[Krueger et~al.(2020)Krueger, Caballero, Jacobsen, Zhang, Binas, Priol,
  and Courville]{krueger2020out}
David Krueger, Ethan Caballero, Joern-Henrik Jacobsen, Amy Zhang, Jonathan
  Binas, Remi~Le Priol, and Aaron Courville.
\newblock Out-of-distribution generalization via risk extrapolation (rex).
\newblock \emph{arXiv preprint arXiv:2003.00688}, 2020.

\bibitem[Parascandolo et~al.(2020)Parascandolo, Neitz, Orvieto, Gresele, and
  Sch{\"o}lkopf]{parascandolo2020learning}
Giambattista Parascandolo, Alexander Neitz, Antonio Orvieto, Luigi Gresele, and
  Bernhard Sch{\"o}lkopf.
\newblock Learning explanations that are hard to vary.
\newblock \emph{arXiv preprint arXiv:2009.00329}, 2020.

\bibitem[Pearl(2009)]{pearl2009causality}
Judea Pearl.
\newblock \emph{Causality}.
\newblock Cambridge university press, 2009.

\bibitem[Rosenfeld et~al.(2021)Rosenfeld, Ravikumar, and
  Risteski]{rosenfeld2021the}
Elan Rosenfeld, Pradeep~Kumar Ravikumar, and Andrej Risteski.
\newblock The risks of invariant risk minimization.
\newblock In \emph{International Conference on Learning Representations}, 2021.
\newblock URL \url{https://openreview.net/forum?id=BbNIbVPJ-42}.

\bibitem[Shi et~al.(2020)Shi, Veitch, and Blei]{shi2020invariant}
Claudia Shi, Victor Veitch, and David Blei.
\newblock Invariant representation learning for treatment effect estimation.
\newblock \emph{arXiv preprint arXiv:2011.12379}, 2020.

\bibitem[Teney et~al.(2020)Teney, Abbasnejad, and Hengel]{teney2020unshuffling}
Damien Teney, Ehsan Abbasnejad, and Anton van~den Hengel.
\newblock Unshuffling data for improved generalization.
\newblock \emph{arXiv preprint arXiv:2002.11894}, 2020.

\bibitem[Vapnik(1992)]{vapnik1992principles}
Vladimir Vapnik.
\newblock Principles of risk minimization for learning theory.
\newblock In \emph{Advances in neural information processing systems}, pages
  831--838, 1992.

\end{thebibliography}

\appendix
\section{Mathematical Proofs} \label{sec:proofs}
\subsection{Proof of Results in Section 2}
\begin{lemm} %\label{lem:risk diff}
Consider squared loss function. Let $w\in \RR^{d_\varphi}$ and $w_e^\star(\varphi)$ as defined in Equation \eqref{eq:LSE}.  Then, 
\begin{align}
     &R_e\left(w^\top \varphi\right) = R_e\left(w_e^\star(\varphi)^\top \varphi\right) + \left\| \bphi{e}{}^{1/2} \left(w - w_e^\star(\varphi)\right)  \right\|^2. %\label{eq:risk diff}
\end{align}
\end{lemm}
\emph{Proof of Lemma \ref{lem:risk diff}:} First, the risk under environment $e$ with $w=w_e^\star(\varphi)$ is given by
\begin{align*}
    &R_e\left(w_e^\star(\varphi)^\top \varphi\right) = \bE_{Y^e}\left[\|Y^e\|^2\right]  
    -\bE_{X^e,Y^e}\left[\varphi(X^e) Y^e \right]^\top \bphi{e}{}^{-1} \bE_{X^e,Y^e}\left[\varphi(X^e) Y^e \right].
\end{align*}
Then,
\begin{align*}
  &R_e\left(w^\top \varphi\right) - R_e\left(w_e^\star(\varphi)^\top \varphi\right)\nonumber\\
  &=w^\top \bphi{e}{} w  - 2w^\top  \bE_{X^e,Y^e}\left[\varphi(X^e) Y^e \right] -\bE_{X^e,Y^e}\left[\varphi(X^e) Y^e \right]^\top \bphi{e}{}^{-1} \bE_{X^e,Y^e}\left[\varphi(X^e) Y^e \right]\\
  &=  \left\| \bphi{e}{}^{1/2} \left(w - w_e^\star(\varphi)\right)  \right\|^2.
\end{align*}
\hfill $\blacksquare$

\begin{lemm} %\label{lem:opt class}
Consider squared loss function and fixed $\varphi$. Let $w_e^\star(\varphi)$ and $w^\star(\varphi)$ as defined in Equations \eqref{eq:LSE} and \eqref{eq:w_opt}, respectively. Then,
\begin{align}
    w^\star(\varphi) = \left(\sum_{e\in \cE_{\mathrm{tr}}} \bphi{e}{}\right)^{-1}\left(\sum_{e\in \cE_{\mathrm{tr}}} \bphi{e}{}w_e^\star(\varphi)\right). %\label{eq:w*}
\end{align}
Moreover, 
\begin{align}
     \argmin_w \ \sum_{e\in \cE_{\mathrm{tr}}} R_e\left(w^\top \varphi\right)=w^\star(\varphi). %\label{eq:w_erm}
\end{align}
\end{lemm}
\emph{Proof of Lemma \ref{lem:risk diff}:} Recall the definition of $w^\star(\varphi)$ 
\begin{align*}
    w^\star(\varphi) = \argmin_w \ \sum_{e\in \cE_{\mathrm{tr}}} R_e\left(w^\top \varphi\right) + \lambda \rho_e^{\mathrm{IRMv2}}(\varphi, w).
\end{align*}
That is,
\begin{align*}
    w^\star(\varphi) &= \argmin_w \ \sum_{e\in \cE_{\mathrm{tr}}} \bE\left[\left\|w^\top \varphi(X^e) -Y^e\right\|^2\right] + \lambda \left\| \bphi{e}{}^{1/2} \left(w - w_e^\star(\varphi)\right) \right\|^2\\
    &= \argmin_w \ \sum_{e\in \cE_{\mathrm{tr}}} (1+\lambda) w^\top \bphi{e}{} w - 2w^\top \left( \bE\left[\varphi(X^e)Y^e\right] + \lambda \bphi{e}{} w_e^\star(\varphi) \right)\\
    &\qquad + w^\star(\varphi)^\top\bphi{e}{}w^\star(\varphi) + \bE\left[\|Y^e\|^2\right].
\end{align*}
Note that the objective function is a convex quadratic function of $w$. Hence, using the first-order optimality condition we have that
\begin{align*}
     (1+\lambda) \left(\sum_{e\in \cE_{\mathrm{tr}}}\bphi{e}{}\right)w^\star(\varphi) - \sum_{e\in \cE_{\mathrm{tr}}}\left( \bE\left[\varphi(X^e)Y^e\right] + \lambda \bphi{e}{} w_e^\star(\varphi) \right) =0.
\end{align*}
Then,
\begin{align*}
    w^\star(\varphi) = \frac{1}{1+\lambda} \left(\sum_{e\in \cE_{\mathrm{tr}}}\bphi{e}{}\right)^{-1} \left(\sum_{e\in \cE_{\mathrm{tr}}}\left( \bE\left[\varphi(X^e)Y^e\right] + \lambda \bphi{e}{} w_e^\star(\varphi) \right)\right).
\end{align*}
Recall that $w_e^\star(\varphi) = \bphi{e}{}^{-1}\left[\varphi(X^e)Y^e\right] $. Then, $\left[\varphi(X^e)Y^e\right] = \bphi{e}{} w_e^\star(\varphi) $. Thus,
\begin{align*}
     w^\star(\varphi) = \left(\sum_{e\in \cE_{\mathrm{tr}}}\bphi{e}{}\right)^{-1} \left(\sum_{e\in \cE_{\mathrm{tr}}}\bphi{e}{} w_e^\star(\varphi) \right).
\end{align*}
Finally, using a similar argument, we get
\begin{align*}
    \argmin_w \ \sum_{e\in \cE_{\mathrm{tr}}} R_e\left(w^\top \varphi\right) &= \argmin_w \ \sum_{e\in \cE_{\mathrm{tr}}} \bE\left[\left\|w^\top \varphi(X^e) -Y^e\right\|^2\right] \\
    &= \argmin_w \ \sum_{e\in \cE_{\mathrm{tr}}} w^\top \bphi{e}{} w - 2w^\top \bE\left[\varphi(X^e)Y^e\right] + \bE\left[\|Y^e\|^2\right]\\
    &= \left(\sum_{e\in \cE_{\mathrm{tr}}}\bphi{e}{}\right)^{-1} \left(\sum_{e\in \cE_{\mathrm{tr}}}\bphi{e}{} w_e^\star(\varphi) \right).
\end{align*}
\hfill $\blacksquare$

\begin{lemm} %\label{lem:adaptive irm}
Let $\rho_e^{\mathrm{IRMv1}}(\varphi, w)$ and $\rho_e^{\mathrm{IRMv2}}(\varphi, w)$ be the invariance penalties of the IRMv1 and IRMv2. Then,
\begin{align}
    \lambda_{\min}\left(\mathcal{I}_e(\varphi)\right)\rho_e^{\mathrm{IRMv2}}(\varphi, w) \leq \rho_e^{\mathrm{IRMv1}}\left(\varphi, w\right) \leq \lambda_{\max}\left(\mathcal{I}_e(\varphi)\right)\rho_e^{\mathrm{IRMv2}}(\varphi, w).
\end{align}
\end{lemm}
\emph{Proof of Lemma \ref{lem:risk diff}:} 
For a symmetric matrix $A\in \RR^{d\times d}$ and a vector $u\in \RR^d$, it holds that $ \lambda_{\min}(A) \|u\|^2 \leq  u^\top A u\leq \lambda_{\max}(A) \|u\|^2$. Let $u = \bphi{e}{}^{1/2} \left(w - w_e^\star(\varphi)\right) $ and $A=\bphi{e}{} $. Then,
\begin{align*}
   \left\|\bphi{e}{} \left(w - w_e^\star(\varphi)\right)\right\|^2 &\leq \lambda_{\max}(\bphi{e}{})\left\|\bphi{e}{}^{1/2} \left(w - w_e^\star(\varphi)\right)  \right\|^2\\
  \left\|\bphi{e}{} \left(w - w_e^\star(\varphi)\right)\right\|^2 &\geq  \lambda_{\min}(\bphi{e}{})\left\|\bphi{e}{}^{1/2} \left(w - w_e^\star(\varphi)\right)  \right\|^2. 
\end{align*}
\hfill $\blacksquare$

\subsection{Proof of Results in Section 4}
In order to prove Theorem \ref{thm:lin}, we first find the optimal classifier for a given data representation for the linear setting in the following Lemma.
\begin{lemm} \label{lem:w opt indicator}
Let $w_e^\star(\Phi) = \argmin_w\ R_e(w^\top \Phi)$ where $R_e(w^\top \Phi)$ is defined as
\begin{align*}
    R_e(w^\top \Phi) = \bE\left[ \mbox{tr}\left(\left( w^\top \Phi X^e - \begin{bmatrix}\mathds{1}_{ \{Y=+1\} }\\ \mathds{1}_{ \{Y=-1\} } \end{bmatrix} \right)\left( w^\top \Phi X^e - \begin{bmatrix}\mathds{1}_{ \{Y=+1\} }\\ \mathds{1}_{ \{Y=-1\} } \end{bmatrix} \right)^\top \right)\right].
\end{align*}
Then, 
\begin{align*}
    w_e^\star(\Phi) = \begin{bmatrix} \eta \beta_e^\star(\Phi)&  (1-\eta) \beta_e^\star(\Phi) \end{bmatrix},
\end{align*}
where
\begin{align}
    \beta_e^\star(\Phi) = \frac{1}{1+ \bar{\mu}_e^\top\bar{\Sigma}_e^{-1}\bar{\mu}_e}\bar{\Sigma}_e^{-1}\bar{\mu}_e. \label{eq:beta}
\end{align}
Here, $\bar{\Sigma}_e$ and $\bar{\mu}_e$ are defined as
\begin{align*}
    \Bar{\Sigma}_e &:= \Phi S  \begin{bmatrix}\sigma_c^2I & 0\\0 &\sigma_e^2 I\end{bmatrix} S^\top \Phi^\top,\\
    \bar{\mu}_e &:= \Phi S \begin{bmatrix} \mu_c\\ \mu_e  \end{bmatrix}.
\end{align*}
\end{lemm}
\emph{Proof of Lemma \ref{lem:w opt indicator}:} 
We have that
\begin{align*}
    w_e^\star(\Phi) = \mathcal{I}_e(\Phi)^{-1} \bE_{X^e,Y^e} \left[\Phi X^e {\Tilde{Y}^{e\top}}\right].
\end{align*}
First, we have that
\begin{align*}
    \bE_{X^e,Y^e} \left[\Phi X^e {\Tilde{Y}^{e\top}}\right] &= \Phi S \bE_{X^e,Y^e} \left[\begin{bmatrix}Z_c\\Z_e\end{bmatrix} \begin{bmatrix}\mathds{1}\{Y=1\}\\\mathds{1}\{Y=-1\}\end{bmatrix}^\top \right]\\
    &= \Phi S \begin{bmatrix} \eta \mu_c & (1-\eta)\mu_c\\ \eta\mu_e & (1-\eta)\mu_e \end{bmatrix}\\
    &= \begin{bmatrix} \eta \bar{\mu}_e & (1-\eta) \bar{\mu}_e\end{bmatrix}.
\end{align*}
Thus,
\begin{align}
     w_e^\star(\Phi) = \begin{bmatrix} \eta \beta_e^\star(\Phi) & (1-\eta) \beta_e^\star(\Phi)\end{bmatrix},
\end{align}
where
\begin{align}
    \beta_e^\star(\Phi) = \mathcal{I}_e(\Phi)^{-1} \bar{\mu}_e. \label{eq:beta I}
\end{align}
We now compute $\mathcal{I}_e(\Phi)^{-1}$ in terms of $\bar{\Sigma}_e^{-1}$ and $\bar{\mu}_e$. We have that
\begin{align*}
    \mathcal{I}_e(\Phi) &= \bE_{X^e} \left[\Phi X^e {X^e}^\top \Phi^\top \right]= \Phi S \bE \left[ \begin{bmatrix}Z_c\\Z_e\end{bmatrix}\begin{bmatrix}Z_c\\Z_e\end{bmatrix}^\top \right] S^\top \Phi^\top.
\end{align*}
From the definition of $Z_c$ and $Z_e$, it follows that
\begin{align*}
    \bE \left[ \begin{bmatrix}Z_c\\Z_e\end{bmatrix}\begin{bmatrix}Z_c\\Z_e\end{bmatrix}^\top \right] = \begin{bmatrix}\sigma_c^2I & 0\\0 &\sigma_e^2 I\end{bmatrix} + \begin{bmatrix} \mu_c\\ \mu_e  \end{bmatrix} \begin{bmatrix} \mu_c\\ \mu_e  \end{bmatrix}^\top.
\end{align*}
Thus,
\begin{align*}
     \mathcal{I}_e(\Phi) = \bar{\Sigma}_e + \bar{\mu}_e \bar{\mu}_e^\top
\end{align*}
Using Sherman-Morrison formula, we have that
\begin{align}
    \mathcal{I}_e(\Phi)^{-1} = \bar{\Sigma}_e^{-1} - \frac{\bar{\Sigma}_e^{-1} \bar{\mu}_e \bar{\mu}_e^\top \bar{\Sigma}_e^{-1}}{ 1+ \bar{\mu}_e^\top\bar{\Sigma}_e^{-1}\bar{\mu}_e }. \label{eq:sherman}
\end{align}
Finally, using Equation \eqref{eq:beta I} it follows that
\begin{align*}
    \beta_e^\star(\Phi) = \left(\bar{\Sigma}_e^{-1} - \frac{\bar{\Sigma}_e^{-1} \bar{\mu}_e \bar{\mu}_e^\top \bar{\Sigma}_e^{-1}}{ 1+ \bar{\mu}_e^\top\bar{\Sigma}_e^{-1}\bar{\mu}_e }\right) \bar{\mu}_e = \frac{1}{1+ \bar{\mu}_e^\top\bar{\Sigma}_e^{-1}\bar{\mu}_e}\bar{\Sigma}_e^{-1}\bar{\mu}_e.
\end{align*}
\hfill $\blacksquare$

\begin{theore}\label{thm:lin}
Assume that $|\cE_{\mathrm{tr}}|>d_e$. Consider a linear data representation $\Phi X = AZ_c + BZ_e$ and a classifier $w(\Phi)$ on top of $\Phi$ that is invariant, i.e., $w(\Phi)=w_e^\star(\Phi)$ for all $e\in  \cE_{\mathrm{tr}}$. If non-degeneracy conditions Eqs. (\ref{eq:nondeg mu}-\ref{eq:nondeg sig}) hold, then either $w(\Phi)=0$ or $B=0$.
\end{theore}
The proof of Lemma \ref{thm:lin} closely follows from the proof of \citealp[Lemma C.3.]{rosenfeld2021the}. In what follows, we include a proof to keep the manuscript self-contained.

\emph{Proof of Theorem \ref{thm:lin}:} First, notice that the decomposition $\varphi(X^e) = A Z_c+BZ_e$ (or $\Phi S = \begin{bmatrix}A&B\end{bmatrix}$) is without loss of generality under the assumption of left-invertibilibity of $S$. Then,
\begin{align}
    \bar{\Sigma}_e &= \sigma_c^2 AA^\top + \sigma_e^2 BB^\top, \label{eq:sig a}\\
    \bar{\mu}_e &= A\mu_c + B\mu_e. \label{eq:mu a}
\end{align}
Recall from Lemma \ref{lem:w opt indicator} that $\beta_e^\star(\Phi) = \bar{\Sigma}_e^{-1}\bar{\mu}_e/ (1+ \bar{\mu}_e^\top\bar{\Sigma}_e^{-1}\bar{\mu}_e)$. If  $\beta_e^\star(\Phi)$ is invariant, then $\beta^\star =  \beta_e^\star(\Phi)$ for all $e\in \cE_{\mathrm{tr}}$. Then, by reorganizing terms, we get
\begin{align*}
  \bar{\Sigma}_e\beta^\star = \left(1- \bar{\mu}_e^\top \beta^\star \right)\bar{\mu}_e.
\end{align*}
Thus, using Equation \eqref{eq:sig a} and \eqref{eq:mu a}, we have that
\begin{align*}
    (\sigma_c^2 AA^\top + \sigma_e^2 BB^\top)\beta^\star = \left(1- (A\mu_c + B\mu_e)^\top \beta^\star\right) (A\mu_c + B\mu_e).
\end{align*}
Let $v:= -\sigma^2 AA^\top \beta^\star +  (1-\mu_c^\top A \beta^\star)A\mu_c$.  Then,
\begin{align}
     B \left(\sigma_e^2 I + \mu_e \mu_e^\top\right) B^\top\beta^\star - v = \left(1- \mu_c^\top A^\top \beta^\star\right)  B\mu_e + \mu_e^\top B^\top \beta^\star A \mu_c. \label{eq:inv B}
\end{align}
Similar to proof of Lemma C.3. in \citep{rosenfeld2021the}, we show that for all fixed $\beta^\star$ and $A$ Eq. \eqref{eq:inv B} for all environments only holds (with probability 1) if $B=0$. If $ |\cE_{\mathrm{tr}}|>d_e$. Then, by the degeneracy assumption of the training sets, there exists at least one environment for which Eq. \eqref{eq:nondeg mu} holds. Let $\tilde{\mu}$ and $\tilde{\sigma}^2$ be the mean of $Z_e$ and variance of $W_e$ for this environment. Then, we have that $\tilde{\mu} = \sum_{i=1}^{d_e} \alpha_i \mu_i$. By applying this linear combination to Eq. \eqref{eq:inv B} for this environment, we get
\begin{align}
     B \left(\tilde{\sigma}^2 I + \tilde{\mu} \tilde{\mu}^\top\right) B^\top\beta^\star - v &= \left(1- \mu_c^\top A^\top \beta^\star\right)  B\sum_{i=1}^{d_e} \alpha_i \mu_i  +  \left(\sum_{i=1}^{d_e} \alpha_i \mu_i\right)^\top B^\top \beta^\star A \mu_c \nonumber\\
     &=\sum_{i=1}^{d_e} \alpha_i \left(\left(1- \mu_c^\top A^\top \beta^\star\right)  B\mu_i + \mu_i^\top B^\top \beta^\star A \mu_c\right)\nonumber\\
     &= \sum_{i=1}^{d_e} \alpha_i \left( B \left(\sigma_i^2 I + \mu_i \mu_i^\top\right) B^\top\beta^\star - v \right), \label{eq:lin B v}
\end{align}
where in the last identity, we applied Eq. \eqref{eq:inv B} for all $i=1,\ldots, d_e$. By rearranging the terms in Eq. \eqref{eq:lin B v}, we get
\begin{align}
    B \left(\tilde{\sigma}^2 I + \tilde{\mu} \tilde{\mu}^\top- \sum_{i=1}^{d_e} \alpha_i  \left(\sigma_i^2 I + \mu_i \mu_i^\top\right) \right) B^\top\beta^\star = \left(1-\sum_{i=1}^{d_e} \alpha_i\right) v. \label{eq:lin B v 2}
\end{align}
From the non-degeneracy condition \eqref{eq:nondeg alpha}, Eq. \eqref{eq:lin B v 2} is equivalent to
\begin{align}
     B \Gamma_{\alpha} B^\top\beta^\star =  v, \label{eq:Gamma cond}
\end{align}
where $\Gamma_{\alpha}$ is defined as
\begin{align*}
   \Gamma_{\alpha}:=\frac{\tilde{\sigma}^2 I + \tilde{\mu} \tilde{\mu}^\top- \sum_{i=1}^{d_e} \alpha_i  \left(\sigma_i^2 I + \mu_i \mu_i^\top\right)}{1-\sum_{i=1}^{d_e} \alpha_i}.
\end{align*}
Note that $B$, $\beta^\star$, and $v$ are environment independent and $\Gamma_{\alpha}$ is an environment dependent matrix for which it holds that $\mbox{rank}(\Gamma_{\alpha})=d_e$ from the nondegeneracy condition \eqref{eq:nondeg sig}. Thus, Eq. \eqref{eq:Gamma cond} holds if and only $v= B \Gamma_{\alpha} B^\top\beta^\star =0$. Then, Eq. \eqref{eq:inv B} reduces to
\begin{align*}
    \left(1- \mu_c^\top A^\top \beta^\star\right)  B\mu_e + {\beta^\star}^\top B\mu_e A \mu_c = 0
\end{align*}
for all $e \in \cE_{\mathrm{tr}}$. Thus, $B\mu_e=0$ for all $e \in \cE_{\mathrm{tr}}$, which holds if and only if $B=0$ given that the span of $\mu_e$s is $\RR^{d_e}$.

\section{The Role of the Eigenstructure of $\bphi{e}{}$} \label{sec:example}
In this section, we elaborate on the discussions on the eigenstructure of $\bphi{e}{}$, and in particular, its condition number in the examples of \citep{arjovsky2019invariant} and \citep{rosenfeld2021the}.
\subsection{Example 1 of \citep{arjovsky2019invariant}} \label{app:arj}
\citet{arjovsky2019invariant} consider data that is generated according to the following SEM
\begin{align*}
    X_1&\sim \cN (0,\sigma^2),\\
    Y  &= X_1 + Z_1,\\
    X_2&= Y + Z_2,
\end{align*}
where $Z_1\sim \cN(0,\sigma^2)$, $Z_2\sim (0,1)$, and $X_1$ are independent, and $X=\begin{bmatrix}X_1\\X_2\end{bmatrix}$. Consider the following data representation
\begin{align}
    \varphi_c(X) = \begin{bmatrix}X_1\\cX_2\end{bmatrix}.
\end{align}
Then,
\begin{align*}
    \mathcal{I}(\varphi_c) = \bE \left[\begin{bmatrix}X_1\\cX_2\end{bmatrix}\begin{bmatrix}X_1\\cX_2\end{bmatrix}^\top \right] = \begin{bmatrix} \sigma^2& c\sigma^2\\
    c\sigma^2& c^2(2\sigma^2+1)\end{bmatrix}.
\end{align*}
We now find the eigenvalues of $\mathcal{I}(\varphi_c)$. That is, the solutions to $\mbox{det}(\mathcal{I}(\varphi_c) -\lambda I ) =0$.
\begin{align*}
    \lambda^2 - \lambda \left(\sigma^2+ c^2(2\sigma^2+1)\right) + c^2\sigma^2 (2\sigma^2+1) - c^2\sigma^4 = 0.
\end{align*}
Then,
\begin{align*}
    \lambda &= \frac{1}{2}\left(\sigma^2+ c^2(2\sigma^2+1) \pm \sqrt{\left(\sigma^2+ c^2(2\sigma^2+1)\right)^2 - 4c^2\sigma^2 (\sigma^2+1)} \right)\\
    &=\frac{1}{2}\left(\sigma^2+ c^2(2\sigma^2+1) \pm \sqrt{\sigma^4 + c^4 (2\sigma^2+1)^2 - 2c^2\sigma^2 } \right).
\end{align*}
Hence,
\begin{align*}
    \kappa (\mathcal{I}(\varphi_c)) &= \frac{\lambda_{\max}(\mathcal{I}(\varphi_c))}{\lambda_{\min}(\mathcal{I}(\varphi_c))}\\
    &= \frac{\sigma^2+ c^2(2\sigma^2+1) +\sqrt{\sigma^4 + c^4 (2\sigma^2+1)^2 - 2c^2\sigma^2 } }{\sigma^2+ c^2(2\sigma^2+1) - \sqrt{\sigma^4 + c^4 (2\sigma^2+1)^2 - 2c^2\sigma^2 } }\\
    &= \frac{\left(\sigma^2+ c^2(2\sigma^2+1) +\sqrt{\sigma^4 + c^4 (2\sigma^2+1)^2 - 2c^2\sigma^2 }\right)^2 }{\left(\sigma^2+ c^2(2\sigma^2+1)\right)^2 - \left(\sigma^4 + c^4 (2\sigma^2+1)^2 - 2c^2\sigma^2 \right) }\\
    &= \frac{1}{2(\sigma^2+1)} \left(\frac{1}{c}\sigma^2+ c(2\sigma^2+1) +\sqrt{\frac{1}{c^2}\sigma^4 + c^2 (2\sigma^2+1)^2 - 2\sigma^2 }\right)^2.
\end{align*}
Finally,
\begin{align*}
    \lim_{c\rightarrow \infty} \kappa (\mathcal{I}(\varphi_c)) = \lim_{c\rightarrow 0} \kappa (\mathcal{I}(\varphi_c)) = \infty.
\end{align*}
It is worth noting that \cite{arjovsky2019invariant} discuss that $\|w-w^\star_e(\varphi) \|^2$ is poor discrepancy both for the invariant data representation, i.e., $c=0$, and for a data representation that heavily rely on the spurious features, i.e., large $c$.

\subsection{Example of \citep{rosenfeld2021the}} \label{app:rosen}
Recall that
\begin{align*}
    \varphi_\epsilon(X^e)=\begin{bmatrix}Z_c\\0\end{bmatrix} \mathds{1}_{ \{Z_e \in \mathcal{Z}_{\epsilon}\} }+ \begin{bmatrix}Z_c\\Z_e\end{bmatrix} \mathds{1}_{ \{Z_e \notin \mathcal{Z}_{\epsilon}\} }.
\end{align*}
Here, $\mathcal{Z}_{\epsilon}$ is defined as
\begin{align}
    \mathcal{Z}_{\epsilon}:= \bigcup_{e\in \cE} \left(\mathcal{B}_r(\mu_e) \cup \mathcal{B}_r(-\mu_e) \right),
\end{align}
where $r:=\sqrt{\epsilon \sigma_e^2d_e}$ and $\mathcal{B}_r(\mu)$ denotes the $\ell-2$ ball of radius $r$ centered at $\mu$. Then,
\begin{align*}
    \bphi{e}{\epsilon} &= \bE\left[\varphi_\epsilon(X^e)\varphi_\epsilon(X^e)^\top\right] = I_c + I_e.
\end{align*}
where $I_c$ and $I_e$ are defined as
\begin{align*}
    I_c &:= \begin{bmatrix}\bE\left[Z_c Z_c^\top | Z_e \in \mathcal{Z}_{\epsilon}\right] &0\\0&0\end{bmatrix} \mathbf{P}\left(Z_e \in \mathcal{Z}_{\epsilon}\right),\\
    I_e &:= \bE \left[\begin{bmatrix}Z_c\\ Z_e\end{bmatrix}\begin{bmatrix}Z_c\\ Z_e\end{bmatrix}^\top \Big| Z_e \in \mathcal{Z}_{\epsilon} \right]\mathbf{P}\left(Z_e \notin \mathcal{Z}_{\epsilon}\right).
\end{align*}
Here, we establish a lower bound on the condition number of $\bphi{e}{\epsilon}$ in terms of the probability of event $\mathds{1}_{ \{Z_e \notin \mathcal{Z}_{\epsilon}\} }$. Using Weyl's inequality, we have that
\begin{align*}
    \lambda_{\max}(\bphi{e}{\epsilon}) &\geq \lambda_{\max}(I_c) + \lambda_{\min}(I_e),\\
    \lambda_{\min}(\bphi{e}{\epsilon}) &\leq \lambda_{\min}(I_c) + \lambda_{\max}(I_e).
\end{align*} 
As $I_e$ is positive semidefinite, $\lambda_{\min}(I_e)\geq 0$. Moreover, $\lambda_{\min}(I_c)=0$. Then,
\begin{align*}
    \lambda_{\max}(\bphi{e}{\epsilon}) &\geq \lambda_{\max}(I_c), \\
    \lambda_{\min}(\bphi{e}{\epsilon}) &\leq \lambda_{\max}(I_e).
\end{align*}
For the first term, we have
\begin{align}
    \lambda_{\max}(I_c) &\geq \frac{1}{d_e+d_c} \mbox{tr}(I_c)\nonumber \\
    &= \frac{1}{d_e+d_c} \bE\left[ \|Z_c\|^2 | Z_e \in \mathcal{Z}_{\epsilon}\right]\mathbf{P}\left(Z_e \in \mathcal{Z}_{\epsilon}\right)\nonumber\\
    &= \frac{1}{d_e+d_c} \bE\left[ \|\mu_c\|^2 Y^2 + \|W_c\|^2 + 2 \mu_c^\top W_c Y | Z_e \in \mathcal{Z}_{\epsilon}\right]\mathbf{P}\left(Z_e \in \mathcal{Z}_{\epsilon}\right)\nonumber\\
    &= \frac{1}{d_e+d_c} (\|\mu_c\|^2 + d_c \sigma_c^2)\mathbf{P}\left(Z_e \in \mathcal{Z}_{\epsilon}\right), \label{eq:max I_c}
\end{align}
where the last identity follows from the fact that $Y^2=1$ almost surely, and the fact that $W_c$ is independent of $Y$ and $W_e$, and hence is independent of the event $\mathds{1}_{ \{Z_e \notin \mathcal{Z}_{\epsilon}\} }$. 

For the second term, we have
\begin{align*}
    \lambda_{\max}(I_e) &\leq \mbox{tr}(I_c)\\
    &=\bE\left[ \|Z_c\|^2 + \|Z_e\|^2 | Z_e \notin \mathcal{Z}_{\epsilon}\right]\mathbf{P}\left(Z_e \notin \mathcal{Z}_{\epsilon}\right)\\
    &= \left( \|\mu_c\|^2 + d_c \sigma_c^2 + \bE\left[ \|Z_e\|^2 | Z_e \notin \mathcal{Z}_{\epsilon}\right] \right) \mathbf{P}\left(Z_e \notin \mathcal{Z}_{\epsilon}\right),
\end{align*}
where the last identity follows similarly as Equation \eqref{eq:max I_c}. Then,
\begin{align*}
    \bE\left[ \|Z_e\|^2 | Z_e \notin \mathcal{Z}_{\epsilon}\right] &= \bE\left[ \|\mu_e\|^2 Y^2 + \|W_e\|^2+ 2 \mu_e^\top W_e Y | Z_e \notin \mathcal{Z}_{\epsilon}\right]\\
    &= \|\mu_e\|^2 + \bE\left[ \|W_e\|^2+ 2 \mu_e^\top W_e Y | Z_e \notin \mathcal{Z}_{\epsilon}\right].
\end{align*}
We have that 
\begin{align*}
    \bE\left[ \|W_e\|^2 | Z_e \notin \mathcal{Z}_{\epsilon}\right] \leq d_e \sigma_e^2.
\end{align*}
Moreover, using Cauchy-Schwarz inequality, we get
\begin{align*}
    \bE\left[  \mu_e^\top W_e Y | Z_e \notin \mathcal{Z}_{\epsilon}\right] \leq \|\mu_e\|\bE\left[  \|W_e\| |Y| | Z_e \notin \mathcal{Z}_{\epsilon}\right] \leq \|\mu_e\| \sqrt{d_e\sigma_e^2}.
\end{align*}
Hence,
\begin{align*}
    \lambda_{\max}(I_e) &\leq \left(\|\mu_c\|^2 + d_c \sigma_c^2 + (\|\mu_e\| + \sqrt{d_e \sigma_e^2})^2\right)\mathbf{P}\left(Z_e \notin \mathcal{Z}_{\epsilon}\right).
\end{align*}
Thus,
\begin{align*}
    \kappa(\bphi{e}{\epsilon}) &\geq \frac{\lambda_{\max}(I_c)}{\lambda_{\max}(I_e)}\\
    &\geq \frac{(\|\mu_c\|^2 + d_c \sigma_c^2)\mathbf{P}\left(Z_e \in \mathcal{Z}_{\epsilon}\right)/(d_e+d_c)}{\left(\|\mu_c\|^2 + d_c \sigma_c^2 + (\|\mu_e\| + \sqrt{d_e \sigma_e^2})^2\right)\mathbf{P}\left(Z_e \notin \mathcal{Z}_{\epsilon}\right)}\\
    &= \frac{\|\mu_c\|^2 + d_c \sigma_c^2}{(d_e+d_c)\left(\|\mu_c\|^2 + d_c \sigma_c^2 + (\|\mu_e\| + \sqrt{d_e \sigma_e^2})^2\right)} \left(\frac{1}{\mathbf{P}\left(Z_e \notin \mathcal{Z}_{\epsilon}\right)} -1\right).
\end{align*}
Note that \cite[Lemma F.3.]{rosenfeld2021the} show that
\begin{align*}
    \mathbf{P}\left(Z_e \notin \mathcal{Z}_{\epsilon}\right)\leq p_{\epsilon,e},
\end{align*}
where
\begin{align}
   p_{\epsilon,e}:= \exp\left(-\frac{1}{8} \min\{\epsilon-1, (\epsilon-1)^2 \} \right)
\end{align}
Then,
\begin{align*}
     \kappa(\bphi{e}{\epsilon}) \geq \frac{\|\mu_c\|^2 + d_c \sigma_c^2}{(d_e+d_c)\left(\|\mu_c\|^2 + d_c \sigma_c^2 + (\|\mu_e\| + \sqrt{d_e \sigma_e^2})^2\right)} \left(\frac{1}{p_{\epsilon,e}} -1\right).
\end{align*}
\cite{rosenfeld2021the} show that the invariance penalty of \citep{arjovsky2019invariant} is no greater than $O(p_{\epsilon,e}^2)$, which can be made arbitrarily small by choosing appropriately large $\epsilon$. However, for such choices of $\epsilon$, matrix $\bphi{e}{\epsilon}$ is ill-conditioned. In particular,
\begin{align*}
    \lim_{\epsilon\rightarrow\infty}\ \kappa(\bphi{e}{\epsilon}) = \infty.
\end{align*}
\section{Full DomainBed results}\label{sec:domainbed}

\subsection{Model selection: training-domain validation set}

\subsubsection{ColoredMNIST}

\begin{center}
\adjustbox{max width=\textwidth}{%
\begin{tabular}{lcccc}
\toprule
\textbf{Algorithm}   & \textbf{+90\%}       & \textbf{+80\%}       & \textbf{-90\%}       & \textbf{Avg}         \\
\midrule
ERM                  & 72.8 $\pm$ 0.1       & 72.6 $\pm$ 0.2       & 9.8 $\pm$ 0.0        & 51.7                 \\
IRMv1                  & 72.5 $\pm$ 0.3       & 72.9 $\pm$ 0.1       & 9.9 $\pm$ 0.1        & 51.8                 \\
IRMv1A                 & 70.7 $\pm$ 0.3       & 72.3 $\pm$ 0.5       & 9.7 $\pm$ 0.0        & 50.9                 \\
IRMv2              & 69.8 $\pm$ 0.8       & 72.9 $\pm$ 0.3       & 9.8 $\pm$ 0.1        & 50.8                 \\
\bottomrule
\end{tabular}}
\end{center}

\subsubsection{RotatedMNIST}

\begin{center}
\adjustbox{max width=\textwidth}{%
\begin{tabular}{lccccccc}
\toprule
\textbf{Algorithm}   & \textbf{0}           & \textbf{15}          & \textbf{30}          & \textbf{45}          & \textbf{60}          & \textbf{75}          & \textbf{Avg}         \\
\midrule
ERM                  & 93.1 $\pm$ 0.1       & 97.8 $\pm$ 0.0       & 98.4 $\pm$ 0.0       & 98.3 $\pm$ 0.1       & 98.2 $\pm$ 0.0       & 94.3 $\pm$ 0.1       & 96.7                 \\
IRMv1                  & 89.6 $\pm$ 2.1       & 96.8 $\pm$ 0.1       & 97.9 $\pm$ 0.1       & 97.8 $\pm$ 0.1       & 97.5 $\pm$ 0.1       & 91.6 $\pm$ 0.0       & 95.2                 \\
IRMv1A                 & 75.9 $\pm$ 5.9       & 71.1 $\pm$ 17.7      & 60.8 $\pm$ 25.1      & 60.4 $\pm$ 25.8      & 60.2 $\pm$ 25.8      & 59.8 $\pm$ 20.5      & 64.7                 \\
IRMv2              & 94.1 $\pm$ 0.0       & 98.2 $\pm$ 0.0       & 98.5 $\pm$ 0.1       & 98.4 $\pm$ 0.1       & 98.3 $\pm$ 0.0       & 95.1 $\pm$ 0.2       & 97.1                 \\
\bottomrule
\end{tabular}}
\end{center}

\subsubsection{PACS}

\begin{center}
\adjustbox{max width=\textwidth}{%
\begin{tabular}{lccccc}
\toprule
\textbf{Algorithm}   & \textbf{A}           & \textbf{C}           & \textbf{P}           & \textbf{S}           & \textbf{Avg}         \\
\midrule
ERM                  & 84.5 $\pm$ 1.6       & 77.1 $\pm$ 0.8       & 96.9 $\pm$ 0.3       & 65.8 $\pm$ 1.9       & 81.1                 \\
IRMv1                  & 77.0 $\pm$ 3.0       & 76.7 $\pm$ 1.1       & 96.4 $\pm$ 0.4       & 64.4 $\pm$ 0.3       & 78.6                 \\
IRMv1A                 & 82.6 $\pm$ 0.5       & 77.7 $\pm$ 0.7       & 96.6 $\pm$ 0.4       & 66.7 $\pm$ 0.5       & 80.9                 \\
IRMv2              & 86.0 $\pm$ 0.9       & 76.6 $\pm$ 0.7       & 96.9 $\pm$ 0.0       & 70.8 $\pm$ 2.0       & 82.6                 \\
\bottomrule
\end{tabular}}
\end{center}

\subsubsection{VLCS}

\begin{center}
\adjustbox{max width=\textwidth}{%
\begin{tabular}{lccccc}
\toprule
\textbf{Algorithm}   & \textbf{C}           & \textbf{L}           & \textbf{S}           & \textbf{V}           & \textbf{Avg}         \\
\midrule
ERM                  & 97.4 $\pm$ 0.1       & 65.0 $\pm$ 0.9       & 74.3 $\pm$ 1.1       & 78.7 $\pm$ 0.1       & 78.8                 \\
IRM                  & 96.3 $\pm$ 0.6       & 61.7 $\pm$ 0.3       & 70.1 $\pm$ 0.1       & 76.0 $\pm$ 1.8       & 76.0                 \\
IRMA                 & 96.9 $\pm$ 0.8       & 64.8 $\pm$ 0.0       & 70.7 $\pm$ 1.4       & 77.0 $\pm$ 0.4       & 77.3                 \\
IRMv2              & 96.6 $\pm$ 1.1       & 65.4 $\pm$ 1.5       & 73.5 $\pm$ 0.5       & 70.6 $\pm$ 2.4       & 76.5                 \\
\bottomrule
\end{tabular}}
\end{center}

\subsubsection{Averages}

\begin{center}
\adjustbox{max width=\textwidth}{%
\begin{tabular}{lccccc}
\toprule
\textbf{Algorithm}        & \textbf{ColoredMNIST}     & \textbf{RotatedMNIST}  & \textbf{PACS} & \textbf{VLCS}    & \textbf{Avg}              \\
\midrule
ERM                       & 51.7 $\pm$ 0.1            & 96.7 $\pm$ 0.0     & 81.1 $\pm$ 0.1  &    78.8 $\pm$ 0.4     & 77.0                      \\
IRMv1                       & 51.8 $\pm$ 0.2            & 95.2 $\pm$ 0.4  & 78.6 $\pm$ 1.0   &  76.0 $\pm$ 0.5        & 75.4                      \\
IRMv1A                      & 50.9 $\pm$ 0.1            & 64.7 $\pm$ 20.1     &  80.9 $\pm$ 0.0 & 77.3 $\pm$ 0.2      & 68.4                     \\
IRMv2                   & 50.8 $\pm$ 0.4            & 97.1 $\pm$ 0.0     &  82.6 $\pm$ 0.9 &    76.5 $\pm$ 0.4     & 76.8                      \\
\bottomrule
\end{tabular}}
\end{center}

\subsection{Model selection: leave-one-domain-out cross-validation}

\subsubsection{ColoredMNIST}

\begin{center}
\adjustbox{max width=\textwidth}{%
\begin{tabular}{lcccc}
\toprule
\textbf{Algorithm}   & \textbf{+90\%}       & \textbf{+80\%}       & \textbf{-90\%}       & \textbf{Avg}         \\
\midrule
ERM                  & 30.4 $\pm$ 13.4      & 50.5 $\pm$ 0.6       & 9.9 $\pm$ 0.0        & 30.2                 \\
IRMv1                  & 50.1 $\pm$ 0.4       & 60.6 $\pm$ 7.3       & 30.0 $\pm$ 14.1      & 46.9                 \\
IRMv1A                 & 69.5 $\pm$ 14.4      & 49.8 $\pm$ 0.2       & 10.0 $\pm$ 0.1       & 43.1                 \\
IRMv2              & 10.0 $\pm$ 0.1       & 36.4 $\pm$ 3.0       & 9.9 $\pm$ 0.0        & 18.8                 \\
\bottomrule
\end{tabular}}
\end{center}

\subsubsection{RotatedMNIST}

\begin{center}
\adjustbox{max width=\textwidth}{%
\begin{tabular}{lccccccc}
\toprule
\textbf{Algorithm}   & \textbf{0}           & \textbf{15}          & \textbf{30}          & \textbf{45}          & \textbf{60}          & \textbf{75}          & \textbf{Avg}         \\
\midrule
ERM                  & 90.3 $\pm$ 1.8       & 97.8 $\pm$ 0.0       & 98.2 $\pm$ 0.1       & 98.2 $\pm$ 0.1       & 97.8 $\pm$ 0.2       & 93.5 $\pm$ 0.4       & 96.0                 \\
IRMv1                  & 89.6 $\pm$ 2.1       & 96.0 $\pm$ 0.5       & 97.9 $\pm$ 0.1       & 97.2 $\pm$ 0.0       & 97.0 $\pm$ 0.2       & 90.9 $\pm$ 0.5       & 94.8                 \\
IRMv1A                 & 75.9 $\pm$ 5.9       & 64.2 $\pm$ 22.4      & 59.4 $\pm$ 26.1      & 59.8 $\pm$ 26.3      & 59.0 $\pm$ 26.7      & 55.9 $\pm$ 22.5      & 62.4                 \\
IRMv2              & 94.1 $\pm$ 0.0       & 98.1 $\pm$ 0.3       & 98.5 $\pm$ 0.1       & 98.2 $\pm$ 0.1       & 98.3 $\pm$ 0.0       & 94.4 $\pm$ 0.3       & 97.0                 \\
\bottomrule
\end{tabular}}
\end{center}

\subsubsection{PACS}

\begin{center}
\adjustbox{max width=\textwidth}{%
\begin{tabular}{lccccc}
\toprule
\textbf{Algorithm}   & \textbf{A}           & \textbf{C}           & \textbf{P}           & \textbf{S}           & \textbf{Avg}         \\
\midrule
ERM                  & 79.7 $\pm$ 0.6       & 73.0 $\pm$ 3.8       & 97.1 $\pm$ 0.9       & 62.1 $\pm$ 1.5       & 78.0                 \\
IRMv1                  & 67.4 $\pm$ 6.7       & 72.3 $\pm$ 4.3       & 87.7 $\pm$ 5.7       & 64.1 $\pm$ 4.5       & 72.9                 \\
IRMv1A                 & 78.8 $\pm$ 3.2       & 78.9 $\pm$ 0.9       & 96.0 $\pm$ 0.2       & 42.3 $\pm$ 16.3      & 74.0                 \\
IRMv2              & 86.3 $\pm$ 0.3       & 76.8 $\pm$ 0.5       & 97.0 $\pm$ 0.4       & 69.7 $\pm$ 2.6       & 82.5                 \\
\bottomrule
\end{tabular}}
\end{center}

\subsubsection{VLCS}

\begin{center}
\adjustbox{max width=\textwidth}{%
\begin{tabular}{lccccc}
\toprule
\textbf{Algorithm}   & \textbf{C}           & \textbf{L}           & \textbf{S}           & \textbf{V}           & \textbf{Avg}         \\
\midrule
ERM                  & 97.5 $\pm$ 0.8       & 60.3 $\pm$ 3.2       & 70.1 $\pm$ 4.2       & 75.5 $\pm$ 2.8       & 75.9                 \\
IRM                  & 93.3 $\pm$ 2.7       & 61.8 $\pm$ 0.4       & 72.9 $\pm$ 0.9       & 74.1 $\pm$ 3.1       & 75.5                 \\
IRMA                 & 79.2 $\pm$ 12.8      & 66.8 $\pm$ 1.4       & 68.3 $\pm$ 4.1       & 73.5 $\pm$ 1.3       & 71.9                 \\
IRMv2              & 98.2 $\pm$ 0.1       & 63.0 $\pm$ 0.1       & 74.4 $\pm$ 1.2       & 69.9 $\pm$ 0.2       & 76.4                 \\
\bottomrule
\end{tabular}}
\end{center}

\subsubsection{Averages}

\begin{center}
\adjustbox{max width=\textwidth}{%
\begin{tabular}{lccccc}
\toprule
\textbf{Algorithm}        & \textbf{ColoredMNIST}     & \textbf{RotatedMNIST}  &  \textbf{PACS}  & \textbf{VLCS}   & \textbf{Avg}              \\
\midrule
ERM                       & 30.2 $\pm$ 4.3            & 96.0 $\pm$ 0.4     &   78.0 $\pm$ 0.9  &  75.9 $\pm$ 0.7      &   70.0                    \\
IRMv1                       & 46.9 $\pm$ 2.1            & 94.8 $\pm$ 0.2     &  72.9 $\pm$ 3.0  &  75.5 $\pm$ 1.6      &    72.5                    \\
IRMv1A                      & 43.1 $\pm$ 4.8            & 62.4 $\pm$ 21.6    & 74.0 $\pm$ 5.0   &   71.9 $\pm$ 2.8    &   62.8                    \\
IRMv2                   & 18.8 $\pm$ 1.1            & 97.0 $\pm$ 0.1      & 82.5 $\pm$ 1.0    &  76.4 $\pm$ 0.3    &        68.7                \\
\bottomrule
\end{tabular}}
\end{center}

\subsection{Model selection: test-domain validation set (oracle)}

\subsubsection{ColoredMNIST}

\begin{center}
\adjustbox{max width=\textwidth}{%
\begin{tabular}{lcccc}
\toprule
\textbf{Algorithm}   & \textbf{+90\%}       & \textbf{+80\%}       & \textbf{-90\%}       & \textbf{Avg}         \\
\midrule
ERM                  & 72.2 $\pm$ 0.3       & 72.6 $\pm$ 0.4       & 12.4 $\pm$ 1.1       & 52.4                 \\
IRMv1                  & 72.5 $\pm$ 0.3       & 72.9 $\pm$ 0.1       & 63.3 $\pm$ 6.6       & 69.6                 \\
IRMv1A                 & 71.4 $\pm$ 0.2       & 72.8 $\pm$ 0.3       & 50.2 $\pm$ 0.1       & 64.8                 \\
IRMv2              & 70.6 $\pm$ 1.2       & 71.9 $\pm$ 0.6       & 20.9 $\pm$ 0.9       & 54.4                 \\
\bottomrule
\end{tabular}}
\end{center}

\subsubsection{RotatedMNIST}

\begin{center}
\adjustbox{max width=\textwidth}{%
\begin{tabular}{lccccccc}
\toprule
\textbf{Algorithm}   & \textbf{0}           & \textbf{15}          & \textbf{30}          & \textbf{45}          & \textbf{60}          & \textbf{75}          & \textbf{Avg}         \\
\midrule
ERM                  & 92.5 $\pm$ 0.6       & 97.8 $\pm$ 0.0       & 97.9 $\pm$ 0.1       & 97.9 $\pm$ 0.2       & 98.3 $\pm$ 0.1       & 94.3 $\pm$ 0.1       & 96.4                 \\
IRMv1                  & 89.6 $\pm$ 2.1       & 96.8 $\pm$ 0.1       & 98.0 $\pm$ 0.2       & 97.5 $\pm$ 0.3       & 97.5 $\pm$ 0.1       & 91.6 $\pm$ 0.0       & 95.2                 \\
IRMv1A                 & 77.9 $\pm$ 7.4       & 71.1 $\pm$ 17.7      & 59.4 $\pm$ 26.1      & 59.8 $\pm$ 26.3      & 59.0 $\pm$ 26.7      & 55.9 $\pm$ 22.5      & 63.9                 \\
IRMv2              & 94.7 $\pm$ 0.4       & 98.0 $\pm$ 0.2       & 98.5 $\pm$ 0.1       & 98.3 $\pm$ 0.0       & 98.3 $\pm$ 0.0       & 95.0 $\pm$ 0.2       & 97.2                 \\
\bottomrule
\end{tabular}}
\end{center}

\subsubsection{PACS}

\begin{center}
\adjustbox{max width=\textwidth}{%
\begin{tabular}{lccccc}
\toprule
\textbf{Algorithm}   & \textbf{A}           & \textbf{C}           & \textbf{P}           & \textbf{S}           & \textbf{Avg}         \\
\midrule
ERM                  & 83.7 $\pm$ 0.5       & 82.1 $\pm$ 0.2       & 97.5 $\pm$ 0.2       & 69.1 $\pm$ 0.6       & 83.1                 \\
IRMv1                  & 66.7 $\pm$ 4.3       & 68.5 $\pm$ 1.6       & 87.1 $\pm$ 5.3       & 67.7 $\pm$ 2.0       & 72.5                 \\
IRMv1A                 & 83.5 $\pm$ 0.2       & 75.7 $\pm$ 3.2       & 96.4 $\pm$ 0.3       & 68.6 $\pm$ 1.8       & 81.0                 \\
IRMv2              & 84.3 $\pm$ 0.3       & 76.5 $\pm$ 0.7       & 96.8 $\pm$ 0.1       & 70.3 $\pm$ 2.4       & 82.0                 \\
\bottomrule
\end{tabular}}
\end{center}

\subsubsection{VLCS}

\begin{center}
\adjustbox{max width=\textwidth}{%
\begin{tabular}{lccccc}
\toprule
\textbf{Algorithm}   & \textbf{C}           & \textbf{L}           & \textbf{S}           & \textbf{V}           & \textbf{Avg}         \\
\midrule
ERM                  & 98.4 $\pm$ 0.1       & 65.1 $\pm$ 1.4       & 72.9 $\pm$ 2.2       & 77.1 $\pm$ 1.7       & 78.4                 \\
IRM                  & 97.6 $\pm$ 1.2       & 61.9 $\pm$ 0.6       & 62.9 $\pm$ 1.3       & 73.0 $\pm$ 0.3       & 73.9                 \\
IRMA                 & 98.0 $\pm$ 0.1       & 64.9 $\pm$ 1.1       & 71.8 $\pm$ 0.8       & 73.9 $\pm$ 1.5       & 77.1                 \\
IRMv2              & 96.3 $\pm$ 1.0       & 67.1 $\pm$ 0.1       & 70.9 $\pm$ 1.3       & 71.9 $\pm$ 1.5       & 76.5                 \\
\bottomrule
\end{tabular}}
\end{center}

\subsubsection{Averages}
\begin{center}
\adjustbox{max width=\textwidth}{%
\begin{tabular}{lccccc}
\toprule
\textbf{Algorithm}        & \textbf{ColoredMNIST}     & \textbf{RotatedMNIST} &   \textbf{PACS}    & \textbf{VLCS}    & \textbf{Avg}              \\
\midrule
ERM                       & 52.4 $\pm$ 0.1            & 96.4 $\pm$ 0.1       &  83.1 $\pm$ 0.1      &  78.4 $\pm$ 0.6     &       77.6               \\
IRMv1                       & 69.6 $\pm$ 2.3            & 95.2 $\pm$ 0.4    &   72.5 $\pm$ 2.3     &    73.9 $\pm$ 0.2    &      77.8                 \\
IRMv1A                      & 64.8 $\pm$ 0.2            & 63.9 $\pm$ 21.1   &   81.0 $\pm$ 0.4    &     77.1 $\pm$ 0.3   &         71.7              \\
IRMv2                   & 54.4 $\pm$ 0.9            & 97.2 $\pm$ 0.1    &     82.0 $\pm$ 0.7   &   76.5 $\pm$ 1.0     &      77.5                  \\
\bottomrule
\end{tabular}}
\end{center}

\end{document}